\documentclass[10pt,twocolumn]{article}
\usepackage[margin=0.75in]{geometry}
\usepackage[T1]{fontenc}
\usepackage{lmodern}
\usepackage{amsmath,amssymb}
\usepackage{graphicx}
\usepackage{adjustbox}
\usepackage{booktabs}
\usepackage{array}
\usepackage{siunitx}
\usepackage{xcolor}
\usepackage{xurl}
\usepackage[hidelinks]{hyperref}
\usepackage{cleveref}

\title{Which Depth Estimator for Jetson AGX Orin?\\\large A 2026 Benchmark of Monocular and Stereo Methods}
\author{Leaf Jiang\\NODAR Inc.\\\texttt{leaf@nodarsensor.com}}
\date{}

\begin{document}
\maketitle

\begin{abstract}
The Jetson AGX Orin is a common computer for robots and vehicles, yet measured accuracy, speed and
energy of current depth estimators on it are hard to find. We benchmark nine model families as of
October 2026: the monocular Depth Anything V2 and 3, Depth Pro, MoGe-2, Metric3D v2 and UniDepth V2,
and the stereo methods libSGM, NVIDIA ESS and Fast-FoundationStereo. Every method runs on the Orin as a
TensorRT FP16 engine or native CUDA code at up to three input sizes, up to the full 5.7\,MP, and is
scored for metric accuracy against Middlebury 2014 ground truth (non-occluded pixels of the 10 scenes
whose disparity stays below 256\,px), GPU time, energy and licence. Stereo is far more accurate. At
30\,fps, NVIDIA's ESS reaches 0.70\% AbsRel in 8.6\,ms and libSGM 0.73\% in 25\,ms; with more compute,
Fast-FoundationStereo reaches 0.33\%. The best monocular models reach 12\% with the true focal length
(Depth Anything~3 Metric-L in 33\,ms, Depth Pro in 803\,ms), and 3.8--4.0\% only with an oracle per-image
scale (MoGe-2-L, UniDepth V2-B, Metric3D v2-L). Stereo error falls as the input resolution grows; no
monocular model improves, and the metric scale of most drifts away from the truth outside their training
resolution. At depth discontinuities, which AbsRel barely registers, the monocular models scatter
17--46\% of the edge points between foreground and background, learned stereo 2--9\% and libSGM at full
resolution 1\%. We give recommendations per frame-rate budget and release all code, per-scene results
and engine build scripts.
\end{abstract}

\section{Introduction}
Dense metric depth is a core input for driving, robotics and industrial automation, and the NVIDIA
Jetson AGX Orin is one of the most common computers on which it runs. Two families of camera-based
methods compete for it. \emph{Monocular} depth networks such as Depth Anything
V2~\cite{yang2024depthv2} and 3~\cite{lin2026da3}, Depth Pro~\cite{bochkovskii2025depthpro},
MoGe-2~\cite{wang2025moge2}, Metric3D v2~\cite{hu2024metric3dv2} and UniDepth
V2~\cite{piccinelli2025unidepthv2} infer metric depth from a single image with large vision
transformers. \emph{Stereo} methods triangulate depth from two calibrated cameras, with classical
matching such as semi-global matching (SGM)~\cite{hirschmuller2008sgm} or with learned matchers such as
NVIDIA's ESS~\cite{nvidia2024ess} and Fast-FoundationStereo~\cite{wen2025fastfoundationstereo}.

Choosing among them for an Orin is hard. Papers report accuracy on desktop GPUs, at the networks'
training resolution and often after aligning the prediction to the ground truth; embedded timings, if
any, come from different devices, precisions and input sizes; and licences differ from permissive to
non-commercial. This paper is a practical benchmark for that choice. We run every method on the same
Jetson AGX Orin 64GB as a TensorRT FP16 engine (accuracy from FP32 where FP16 is unreliable) or as native
CUDA code, at up to three input sizes from its native or smallest input to the full 5.7\,MP, and score
every output against the same full-resolution Middlebury 2014 ground truth~\cite{scharstein2014middlebury}
on the 10 scenes whose disparity stays within the 256-level range of the semi-global matcher. We report
metric accuracy, GPU time, energy and licence, and derive recommendations per frame-rate budget.

\paragraph{Contributions.}
(i) Measured metric accuracy, GPU time and energy on Jetson AGX Orin for nine current monocular and
stereo model families, each at up to three input sizes (\cref{tab:main}).
(ii) Recommendations: the most accurate stereo and monocular configuration for each frame-rate budget,
with licences (\cref{sec:recommendations}).
(iii) A resolution-scaling study: stereo error falls with resolution, monocular error does not, because
the monocular metric scale drifts outside the training resolution and the shape does not improve
(\cref{sec:scaling}).
(iv) An analysis of depth discontinuities, where AbsRel is blind to the ``flying'' points that the
networks produce and a classical matcher can be the better choice (\cref{sec:edges}).
(v) All code, per-scene results, engine build scripts and this paper's source, released at
\url{https://github.com/nodarhub/orin-depth-benchmark}.

\section{Related work}
\paragraph{Monocular depth.}
MiDaS~\cite{ranftl2022midas} and DPT~\cite{ranftl2021dpt} established zero-shot relative depth by
mixing datasets. Depth Anything~\cite{yang2024depth,yang2024depthv2} scales this with DINOv2
features~\cite{oquab2024dinov2} and pseudo-labelled data, with metric variants fine-tuned on
Hypersim~\cite{roberts2021hypersim} (indoor) and Virtual KITTI (outdoor). Depth Anything
3~\cite{lin2026da3} predicts depth and camera rays from any number of views with a plain DINOv2
transformer; its monocular metric model predicts depth relative to the focal length. Depth
Pro~\cite{bochkovskii2025depthpro} predicts metric depth and focal length from a multi-scale ViT at
$1536^2$. MoGe-2~\cite{wang2025moge2} predicts an affine-invariant point map, from which it recovers the
focal length and shift, plus a global metric scale from the class token. Metric3D
v2~\cite{hu2024metric3dv2} predicts depth in a canonical camera space with a fixed focal length and
rescales it with the true one; UniDepth V2~\cite{piccinelli2025unidepthv2} predicts a camera
representation together with depth and can take the true intrinsics as input. Tile-based methods such as
PatchFusion~\cite{li2024patchfusion} and PatchRefiner~\cite{li2024patchrefiner} target high-resolution
monocular metric depth by processing overlapping crops and fusing them with a global estimate.

\paragraph{Stereo depth.}
SGM~\cite{hirschmuller2008sgm} remains the workhorse of embedded stereo, with open-source GPU
implementations such as libSGM~\cite{fixstars2024libsgm}. Learned matchers such as
RAFT-Stereo~\cite{lipson2021raft} and FoundationStereo~\cite{wen2025foundationstereo} are far more
accurate but expensive. Fast-FoundationStereo~\cite{wen2025fastfoundationstereo} distils
FoundationStereo into a real-time model. NVIDIA's ESS~\cite{nvidia2024ess}, distributed with Isaac ROS,
is a compact learned stereo network designed for robots on Jetson. Embedded GPUs also ship dedicated
stereo hardware: the Orin's optical flow accelerator (OFA) computes SGM-style disparity, exposed
through NVIDIA VPI; we do not evaluate it.

\paragraph{Monocular versus stereo.}
Pseudo-LiDAR~\cite{wang2019pseudolidar} showed that image-based depth representation matters for
3D detection, with stereo far ahead of monocular. Depth networks are also known to smooth across
depth discontinuities and produce ``flying pixels'' between foreground and background. We are not aware
of a study that compares how the metric accuracy of current monocular foundation models and of stereo
changes with input resolution, or that compares them under a common embedded compute budget.

\section{Methods compared}
\label{sec:methods}
Each method runs once per frame on the full-resolution Middlebury images. Methods that cannot process
the full 5.7\,MP image on the 64\,GB Orin receive the images downsampled to the largest input they
support. Because the monocular networks are trained at much smaller inputs, we also run each of them at
its native input size, and we run libSGM and Fast-FoundationStereo at comparably small inputs so
that both sensor types are compared at low compute as well. Every method that can therefore runs at
two or three input sizes, which gives the resolution-scaling curves of \cref{sec:scaling}: Depth
Anything V2-S and -L, Depth Anything~3, MoGe-2, Metric3D v2 and UniDepth V2 also run at about twice
their native size in each dimension, and libSGM at full, half and quarter resolution. ESS has a fixed
input size.
\Cref{tab:res} lists the processing resolutions, disparity search range and limiting factor of each
method.

\begin{table*}[!t]
\centering
\caption{Configuration of each method. All methods receive the same full-resolution (${\approx}$5.7\,MP)
images and are scored against full-resolution ground truth; methods that cannot process 5.7\,MP on the
64\,GB Jetson AGX Orin run at their maximum input size. Full-resolution equivalents assume a 2880\,px
wide image.}
\label{tab:res}
\small
\begin{tabular}{@{}>{\raggedright\arraybackslash}p{0.14\linewidth}>{\raggedright\arraybackslash}p{0.18\linewidth}>{\raggedright\arraybackslash}p{0.06\linewidth}>{\raggedright\arraybackslash}p{0.18\linewidth}>{\raggedright\arraybackslash}p{0.20\linewidth}>{\raggedright\arraybackslash}p{0.12\linewidth}@{}}
\toprule
Method & Processing resolution & Full res.? & Max.\ disparity search range & Limiting factor & Licence (as stated by the authors) \\
\midrule
\multicolumn{6}{@{}l}{\emph{Stereo}} \\
libSGM (Fixstars, 4 paths) & full ($\approx$1990$\times$2880); also half ($\approx$995$\times$1440) and quarter ($\approx$497$\times$720) & yes & 256 levels at the processing resolution (256, $\approx$512 and $\approx$1024\,px at full resolution) & --- & Apache-2.0 \\
ESS, Light ESS (NVIDIA Isaac ROS) & 576$\times$960 (ESS), 288$\times$480 (Light ESS) & no & no fixed limit (regression) & fixed network input & NVIDIA TAO model EULA (commercial use permitted) \\
Fast-Foundation\-Stereo (both checkpoints) & 768$\times$1152; also 512$\times$768 and 384$\times$576 & no & $0.27\times$ input width: 320, 224 and 160\,px ($\approx$800\,px at full resolution) & does not fit the 64\,GB Orin at larger sizes: no TensorRT implementation for the cost-volume pooling at 960$\times$1440; ONNX export runs out of host memory at 1280$\times$1920 & NVIDIA Source Code License (non-commercial) \\
C-Fast-Foundation\-Stereo (commercial checkpoint) & 768$\times$1152; also 512$\times$768 and 384$\times$576 & no & as Fast-FoundationStereo & as Fast-FoundationStereo & NVIDIA Open Model Agreement \\
\midrule
\multicolumn{6}{@{}l}{\emph{Monocular}} \\
Depth Pro & 1536$\times$1536 & no & n/a & fixed network input & Apple licence \\
Depth Anything V2 (S, B, L) & full (1988$\times$2884); also native 518$\times$770 and (S, L) 1008$\times$1512 & yes & n/a (metric head limited to 20\,m) & --- (trained at 518\,px; FP16 engine of the Small model is numerically unreliable at full size, \cref{sec:fp16}) & S: Apache-2.0; B, L: CC BY-NC 4.0 \\
Depth Anything 3 Metric-L & native 350$\times$504 (longest side 504\,px); also 700$\times$1008 and full (1988$\times$2884) & yes & n/a & --- (trained at 504\,px) & Apache-2.0 \\
MoGe-2 (S, B, L) & native $\approx$700$\times$1008 (3600 tokens, default level); also 1400$\times$2030 and full (1988$\times$2884) & yes & n/a & --- (token count is an input parameter) & MIT \\
Metric3D v2 (ViT-S, ViT-L) & native 616$\times$1064; also 1232$\times$2128 and full (1988$\times$2884) & yes & n/a & --- (input size is a preprocessing parameter) & BSD-2-Clause \\
UniDepth V2 (S, B, L) & native 644$\times$938 (default pixel budget); also 1302$\times$1890 and full (1988$\times$2884) & yes & n/a & --- (pixel budget is a parameter) & CC BY-NC 4.0 \\
\bottomrule
\end{tabular}

\end{table*}

\paragraph{libSGM.} The open-source CUDA semi-global matcher of Fixstars~\cite{fixstars2024libsgm}
(version 3.1.0, Apache-2.0). It computes a per-pixel Hamming cost on symmetric $9\times7$ census codes,
aggregates it along four scan paths (horizontal and vertical), selects the winner with a uniqueness
test and parabolic sub-pixel refinement, applies a $3\times3$ median filter and invalidates pixels that
fail a left--right consistency check. It stores the aggregated cost volume per path with 64-bit
offsets, so it runs at full resolution. We run it on grey-scale input with 256 disparity levels at
full resolution and at half and quarter resolution (area downsampling); the 256 levels cover 255\,px of
full-resolution disparity at full resolution, ${\approx}$510\,px at half and ${\approx}$1020\,px at
quarter resolution. Its smoothness penalties $P_1$ and $P_2$ are tuned separately for each resolution
(\cref{app:libsgm}); all other parameters keep libSGM's defaults.
\paragraph{Fast-FoundationStereo (FFS).} Official research checkpoints \texttt{23-36-37} (8 GRU
iterations, most accurate) and \texttt{20-30-48} (4 iterations, fastest), exported with the
authors' single-ONNX script. The largest input that fits on the 64\,GB Orin is $768\times1152$: at
$960\times1440$ TensorRT finds no implementation for the cost-volume pooling, and at
$1280\times1920$ the ONNX export runs out of host memory. To show how FFS scales, we also run both
checkpoints at $512\times768$ and $384\times576$. The maximum disparity is $0.27\times$ the input width
rounded up to a multiple of 32 (320, 224 and 160\,px; ${\approx}$800\,px at full resolution), which
covers the dataset. We also benchmark C-Fast-FoundationStereo, the commercially licensed checkpoint
(\texttt{nvidia/c-fast-foundationstereo}), at the same three sizes: it has the architecture and
configuration of \texttt{23-36-37} with separately trained weights.
\paragraph{Depth Pro.} Fixed $1536\times1536$ network input. Its metric depth needs a focal length:
we evaluate it both with the \emph{true} focal length from the Middlebury calibration and with Depth
Pro's own field-of-view estimate.
\paragraph{Depth Anything V2 (metric).} The Hypersim (indoor) metric checkpoints, which suit
Middlebury's indoor scenes, at the Small, Base and Large sizes, run at full resolution
($1988\times2884$, i.e.\ $142\times206$ patches of 14\,px) and, for reference, at their native
$518\times770$ input (the 518\,px short side they are trained at) and, for Small and Large, at
$1008\times1512$. Exporting at full resolution
requires memory-efficient attention (identical computation); the TensorRT engines take 27--79\,min to
build and use most of the Orin's 64\,GB.
\paragraph{Depth Anything 3 Metric-Large.} The only monocular metric model of the Depth Anything 3
family (0.35\,B parameters, ViT-L, Apache-2.0). Its official preprocessing resizes the longest side
to 504\,px and rounds both sides to multiples of 14 ($350\times504$, or $336\times504$ for the scenes
with a wider aspect ratio); we also set the longest side to 1008\,px (${\approx}700\times1008$) and run
each image at its own full size, rounded to multiples of 14. Its output becomes metric depth by
multiplying with the focal length in pixels of the processed image and dividing by 300; we use the
true focal length.
\paragraph{MoGe-2.} The Small, Base and Large checkpoints (ViT-S/B/L, the ``-normal'' variants) at
their default resolution level, 3600 patch tokens (${\approx}700\times1008$ for Middlebury's 3:2
images), and at 14{,}400 tokens (${\approx}1400\times2030$) and full resolution (one token per
$14\times14$\,px, ${\approx}$29{,}000 tokens), with the predicted metric scale. Like Depth Pro we evaluate MoGe-2 both with the true focal length, from
which it recovers only the depth shift, and with its own focal-length estimate. We do not apply its
validity mask, so its output is dense like that of the other monocular models.
\paragraph{Metric3D v2.} The ViT-Small and ViT-Large models with the official preprocessing: the image is
resized to fit the $616\times1064$ input keeping its aspect ratio and padded, and the depth predicted in
a canonical camera with a 1000\,px focal length is rescaled with the true focal length of the resized
image. We also use inputs of $1232\times2128$ and $1988\times2884$ (full resolution). Its ViT attention is
computed in blocks of queries at full resolution, which is the same computation in less memory.
\paragraph{UniDepth V2.} The Small, Base and Large models through the official \texttt{infer}, given the
true intrinsics. UniDepth chooses its processing size from a pixel budget (0.2--0.6\,MP by default,
$644\times938$ for Middlebury); we also set budgets that give $1302\times1890$ and full resolution. Its
TensorRT engines take the camera as rays (the official \texttt{UniDepthV2ONNXcam} export).
\paragraph{ESS and Light ESS.} NVIDIA's stereo networks from Isaac ROS (NGC model version 4.1.0), with
fixed inputs of $576\times960$ (ESS) and $288\times480$ (Light ESS); custom layers come from the TensorRT
plugin library distributed with the model. Both images are resized to the network input, which changes
their aspect ratio from 3:2 to 5:3 but not the horizontal disparity, and normalised as on the model card.
We use the dense disparity output, without the optional confidence threshold.

\section{Experimental setup}
\label{sec:setup}
\paragraph{Platform.} NVIDIA Jetson AGX Orin 64GB developer kit, JetPack 6.2 (L4T R36.4.3), CUDA
12.6, TensorRT 10.3, MAXN power mode. All networks are timed as TensorRT FP16 engines built on the
device. Monocular accuracy comes from the FP32 PyTorch models with the official pre- and
post-processing, except for Depth Anything V2 at native input and V2-B/L at full resolution, whose
FP16 engines (and those at $1008\times1512$) agree with FP32 to within 0.25\% and are used for both
(\cref{sec:fp16}). ESS and Light ESS
are evaluated with their TensorRT engines, the only form in which they are distributed.

\paragraph{Data.} Middlebury 2014~\cite{scharstein2014middlebury} with ``perfect'' rectification at full
resolution (${\approx}2.9\text{k}\times2.0\text{k}$). Of the 23 scenes with public ground truth we use
the 10 whose non-occluded ground-truth disparity stays below 256\,px (at most 243\,px), so that the
256-level semi-global matcher is not limited by its disparity range at any of the resolutions we
test; \cref{app:dataset} lists them. In the other 13 scenes near objects reach 265--712\,px of disparity
at full resolution.

\paragraph{Non-occluded ground truth.} Middlebury's ground truth comes from structured light, not from
stereo matching, so it also covers pixels that the right camera cannot see: on average 5.8\% of the
ground-truth pixels are occluded in the right view and 2.7\% fall outside its field of view. No stereo
method can match these pixels, so, following MiddEval3, we evaluate every method only on the
\emph{non-occluded} ground truth (91\% of it): pixels whose match lies inside the right image and
whose right-view ground-truth disparity agrees within 1\,px. Monocular methods are evaluated on the
same pixels.

\paragraph{Protocol.} Downsampled outputs are upsampled to full resolution, bilinearly for the dense
network outputs and with nearest-neighbour interpolation for libSGM, so that its holes stay holes;
disparities are rescaled by the width ratio. Depth follows from disparity via $Z=fB/(d+d_\text{offs})$; monocular
depth is converted to disparity the same way, so a monocular pixel predicted beyond the
zero-disparity distance $fB/d_\text{offs}$ counts as missing.

\paragraph{Metrics.} For disparity: bad-$t$, the percentage of non-occluded ground-truth pixels
with error $>t$\,px, where missing predictions count as errors. For depth~\cite{eigen2014depth}:
AbsRel and the threshold accuracy $\delta<1.05$ (fraction of pixels within 5\% of the true depth),
over the predicted pixels. The uniqueness and left--right checks of libSGM leave holes,
so we report it both raw (density, plus errors on predicted pixels) and after
MiddEval3-style background hole filling (each hole takes the smaller of the nearest valid
disparities to its left and right on the same row). For monocular methods we additionally report
\emph{oracle-scaled} metrics, where each prediction is multiplied by the per-image median ratio to the
non-occluded ground truth, an upper bound on what any scale-recovery scheme could achieve.

\paragraph{Accuracy, not precision.}
All metrics above measure \emph{accuracy}: the error of each depth value with respect to the ground
truth. We deliberately do not report \emph{precision}, i.e.\ the noise or repeatability of the depth
map around its own local or temporal mean, because metric accuracy is what downstream tasks such as
obstacle detection, ground-plane estimation and free-space computation depend on. Precision is
particularly misleading for neural networks. A network can produce a perfectly smooth, noise-free
surface that is nevertheless in the wrong place: in our experience with field data, a monocular
network renders a flat floor as a perfectly flat plane with no visible noise, yet with its tilt off
by several degrees. A precision metric would rate such a hallucinated surface as excellent; an
accuracy metric against independent ground truth exposes the error. The oracle-scaled metrics remove
only a single global scale factor per image, so such structural errors (tilt, warping, misplaced
objects) remain in them as well. Classical stereo shows the opposite pattern: noisier depth values
and missing pixels where matching fails, but on the pixels it does return, its error is governed by
matching and triangulation geometry rather than by a learned prior of what the scene should look
like.

\paragraph{Speed and energy.} For networks we report the median \texttt{trtexec} GPU compute time
(spin-wait, 3\,s warm-up, ${\geq}20$\,s run). For libSGM we report the median
per-frame GPU busy time from an Nsight Systems trace: the union of all kernel and copy intervals,
including the upload of both images and the download of the disparity. Power is
sampled with \texttt{tegrastats} (VDD\_GPU\_SOC + VDD\_CPU\_CV); dynamic energy per frame is
$(P_\text{load}-P_\text{idle})/\text{throughput}$.

\section{Which method to use}
\label{sec:recommendations}

\begin{table*}[!t]
\centering
\caption{Recommendations: the most accurate configuration whose GPU time on one Jetson AGX Orin (MAXN,
FP16) fits the frame-rate budget, by AbsRel on the non-occluded pixels of the 10 Middlebury 2014
scenes. Monocular models with the true focal length or intrinsics; the last column assumes the metric
scale is supplied by an oracle (an upper bound). Last row: the configuration with the lowest GPU time
among the methods whose licences permit commercial use and modification (libSGM, Depth Anything V2-S
and 3, MoGe-2, Metric3D v2). Licences in \cref{tab:res}.}
\label{tab:rec}
\small
\resizebox{\textwidth}{!}{\begin{tabular}{@{}llll@{}}
\toprule
Budget & Best stereo & Best monocular & Best monocular, oracle scale \\
 & AbsRel (GPU time) & AbsRel (GPU time) & AbsRel (GPU time) \\
\midrule
30 fps & ESS (NVIDIA), 576$\times$960: 0.70\% (9\,ms) & Depth Anything 3 Metric-L, 350$\times$504: 12.34\% (33\,ms) & Depth Anything 3 Metric-L, 350$\times$504: 5.97\% (33\,ms) \\
10 fps & Fast-FoundationStereo 20-30-48, 384$\times$576: 0.64\% (68\,ms) & Depth Anything 3 Metric-L, 350$\times$504: 12.34\% (33\,ms) & UniDepth V2-B, 644$\times$938: 3.90\% (85\,ms) \\
1 fps & C-Fast-FoundationStereo, 768$\times$1152: 0.33\% (760\,ms) & Depth Pro, 1536$\times$1536: 11.98\% (803\,ms) & MoGe-2-L, 700$\times$1008: 3.78\% (212\,ms) \\
no limit & C-Fast-FoundationStereo, 768$\times$1152: 0.33\% (760\,ms) & Depth Pro, 1536$\times$1536: 11.98\% (803\,ms) & MoGe-2-L, 700$\times$1008: 3.78\% (212\,ms) \\
\midrule
lowest GPU, open source & libSGM, $\approx$497$\times$720: 1.18\% (6\,ms) & Depth Anything V2-S (native), 518$\times$770: 29.88\% (12\,ms) & Depth Anything V2-S (native), 518$\times$770: 8.93\% (12\,ms) \\
\bottomrule
\end{tabular}
}
\end{table*}

\Cref{tab:rec} summarises the results as recommendations for a single Jetson AGX Orin.

\paragraph{With two cameras.} At 30\,fps, ESS is the best choice: 0.70\% AbsRel with a dense output in
8.6\,ms, and Light ESS 1.16\% in 2.8\,ms, which leaves room for several camera pairs or other networks.
libSGM at half resolution is about as accurate (0.73\% in 25\,ms) on the 85\% of the pixels it returns,
is open source under Apache-2.0 and can be tuned or modified. Where GPU time is the scarcest resource,
libSGM at quarter resolution is the open-source choice: 1.18\% AbsRel on 88\% of the pixels in only
6.1\,ms and 0.18\,J per frame, so that at 30\,fps it occupies 18\% of the Orin's GPU and leaves the rest to
other tasks; only Light ESS needs less (2.8\,ms for 1.16\%, dense). At 10\,fps, Fast-FoundationStereo with 4
iterations at $384\times576$ reaches 0.64\% in 68\,ms, and with more compute 0.33\% (8 iterations,
$768\times1152$, 758\,ms); its research checkpoints are licensed for non-commercial use only. The
commercially licensed C-Fast-FoundationStereo is as accurate as the research checkpoint of the same
architecture (0.33\% at $768\times1152$ in 760\,ms, 0.52\% at $384\times576$ in 116\,ms), so for a
commercial product it is the choice when more than 100\,ms per frame is available.

\paragraph{With one camera.} No monocular model comes close to stereo in metric accuracy. The most
accurate are Depth Anything~3 Metric-Large at its native input (12.3\% in 33\,ms, Apache-2.0) and Depth
Pro (12.0\% in 803\,ms), both given the true focal length. If the metric scale can be obtained from
another source, such as a known object size, a range sensor or odometry, the best shapes come from
MoGe-2-L (3.8\% with oracle scale, 212\,ms, MIT), UniDepth V2-B (3.9\%, 85\,ms, non-commercial) and
Metric3D v2-L (4.0\%, 265\,ms, BSD-2-Clause), and at 30\,fps from Depth Anything~3 (6.0\%). Running the
monocular models above their native input does not help (\cref{sec:scaling}).

\section{Results}

\subsection{Resolution scaling}
\label{sec:scaling}

\begin{figure*}[t]
\centering
\includegraphics[width=\linewidth]{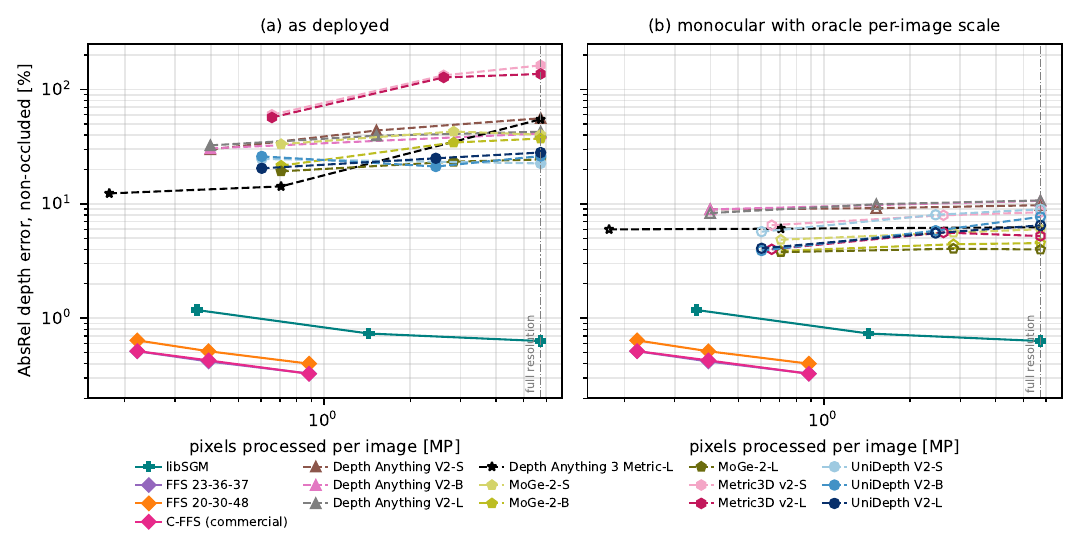}
\caption{Resolution scaling. AbsRel (non-occluded pixels, mean over the 10 scenes, log scale) against
the number of pixels each method processes, for every method run at more than one input size. (a) As
deployed. (b) Monocular models after per-image oracle median scaling (hollow markers), stereo unchanged.
Stereo (solid lines): libSGM at quarter, half and full resolution and Fast-FoundationStereo (FFS) at
$384\times576$, $512\times768$ and $768\times1152$, its largest input on the Orin (ESS has a fixed
input). Monocular (dashed lines): each model from its native input to full resolution, with the true
focal length or intrinsics. Stereo error falls with resolution; monocular error does not, and with an
oracle scale it stays flat or rises.}
\label{fig:scaling}
\end{figure*}

\begin{table}[t]
\centering
\caption{Monocular models across input sizes. Scale: median over the scenes of the per-image median
ratio of ground-truth to predicted depth (1 = unbiased, $>$1 = predicted too close). Density below
100\% means that part of the scene is placed beyond the zero-disparity distance.}
\label{tab:scale}
\small
\resizebox{\columnwidth}{!}{\begin{tabular}{@{}lrrrrrr@{}}
\toprule
Model & Input & MP & Scale GT/pred. & AbsRel [\%] & Oracle AbsRel [\%] & Density [\%] \\
\midrule
Depth Anything V2-S & 518$\times$770 & 0.40 & 1.45 & 29.9 & 8.9 & 100 \\
Depth Anything V2-S & 1008$\times$1512 & 1.52 & 1.88 & 43.8 & 9.2 & 100 \\
Depth Anything V2-S & 1988$\times$2884 & 5.73 & 2.44 & 56.0 & 9.7 & 100 \\
\addlinespace
Depth Anything V2-B & 518$\times$770 & 0.40 & 1.50 & 30.6 & 8.9 & 100 \\
Depth Anything V2-B & 1988$\times$2884 & 5.73 & 1.83 & 41.2 & 10.6 & 100 \\
\addlinespace
Depth Anything V2-L & 518$\times$770 & 0.40 & 1.54 & 32.5 & 8.3 & 100 \\
Depth Anything V2-L & 1008$\times$1512 & 1.52 & 1.78 & 39.5 & 9.9 & 100 \\
Depth Anything V2-L & 1988$\times$2884 & 5.73 & 1.91 & 42.7 & 10.7 & 100 \\
\addlinespace
Depth Anything 3 Metric-L & 350$\times$504 & 0.18 & 0.98 & 12.3 & 6.0 & 93 \\
Depth Anything 3 Metric-L & 700$\times$1008 & 0.71 & 0.93 & 14.2 & 6.1 & 90 \\
Depth Anything 3 Metric-L & 1988$\times$2884 & 5.73 & 0.66 & 54.9 & 6.2 & 56 \\
\addlinespace
MoGe-2-S & 700$\times$1008 & 0.71 & 0.78 & 33.3 & 4.9 & 76 \\
MoGe-2-S & 1400$\times$2030 & 2.84 & 0.73 & 42.9 & 5.6 & 68 \\
MoGe-2-S & 1988$\times$2884 & 5.73 & 0.76 & 40.2 & 6.0 & 66 \\
\addlinespace
MoGe-2-B & 700$\times$1008 & 0.71 & 0.83 & 21.6 & 3.9 & 96 \\
MoGe-2-B & 1400$\times$2030 & 2.84 & 0.76 & 34.4 & 4.4 & 86 \\
MoGe-2-B & 1988$\times$2884 & 5.73 & 0.74 & 37.2 & 4.5 & 81 \\
\addlinespace
MoGe-2-L & 700$\times$1008 & 0.71 & 0.89 & 19.2 & 3.8 & 94 \\
MoGe-2-L & 1400$\times$2030 & 2.84 & 0.85 & 23.4 & 4.1 & 87 \\
MoGe-2-L & 1988$\times$2884 & 5.73 & 0.85 & 24.4 & 4.0 & 86 \\
\addlinespace
Metric3D v2-S & 616$\times$1064 & 0.66 & 0.70 & 60.1 & 6.5 & 65 \\
Metric3D v2-S & 1232$\times$2128 & 2.62 & 0.47 & 133.0 & 7.9 & 14 \\
Metric3D v2-S & 1988$\times$2884 & 5.73 & 0.40 & 162.5 & 8.4 & 5 \\
\addlinespace
Metric3D v2-L & 616$\times$1064 & 0.66 & 0.65 & 56.9 & 4.0 & 63 \\
Metric3D v2-L & 1232$\times$2128 & 2.62 & 0.47 & 127.5 & 5.6 & 9 \\
Metric3D v2-L & 1988$\times$2884 & 5.73 & 0.45 & 136.8 & 5.2 & 8 \\
\addlinespace
UniDepth V2-S & 644$\times$938 & 0.60 & 0.97 & 24.7 & 5.7 & 93 \\
UniDepth V2-S & 1302$\times$1890 & 2.46 & 1.06 & 23.4 & 8.0 & 95 \\
UniDepth V2-S & 1988$\times$2884 & 5.73 & 1.15 & 22.6 & 9.0 & 96 \\
\addlinespace
UniDepth V2-B & 644$\times$938 & 0.60 & 0.91 & 26.0 & 3.9 & 84 \\
UniDepth V2-B & 1302$\times$1890 & 2.46 & 1.15 & 21.2 & 5.8 & 95 \\
UniDepth V2-B & 1988$\times$2884 & 5.73 & 1.37 & 26.2 & 7.7 & 96 \\
\addlinespace
UniDepth V2-L & 644$\times$938 & 0.60 & 1.15 & 20.5 & 4.1 & 99 \\
UniDepth V2-L & 1302$\times$1890 & 2.46 & 1.28 & 25.0 & 5.6 & 100 \\
UniDepth V2-L & 1988$\times$2884 & 5.73 & 1.34 & 28.1 & 6.5 & 100 \\
\bottomrule
\end{tabular}
}
\end{table}

\Cref{fig:scaling} plots AbsRel against the number of pixels each method processes, for every method
that runs at more than one input size. Stereo error falls with every increase in resolution. libSGM
goes from 1.18\% at quarter resolution (0.36\,MP) to 0.73\% at half and 0.63\% at full resolution
(5.7\,MP); Fast-FoundationStereo from 0.51\% to 0.42\% and 0.33\% with 8 iterations and from 0.64\% to
0.51\% and 0.40\% with 4, between 0.22 and 0.88\,MP. The gains are smaller than the ideal of the
triangulation argument, which halves the error with every doubling of the image side at a constant
matching error in pixels; at the last step libSGM gains 14\%. The matching error in pixels grows at
higher resolution, most likely because a fixed-size matching window then covers less of the scene.

No monocular model improves with resolution. Depth Anything~3 goes from 12.3\% at its native
$350\times504$ input to 14.2\% at $700\times1008$ and 54.9\% at full resolution, MoGe-2-L from 19.2\%
to 23.4\% and 24.4\%, Depth Anything V2-L from 32.5\% to 39.5\% and 42.7\%, and V2-S from 29.9\% to
43.8\% and 56.0\%. Metric3D v2 collapses above its native input: its error rises from 57--60\% to
128--162\%, and it places 86--95\% of the pixels beyond the zero-disparity distance. UniDepth V2 stays
at 20--28\% at every size, slightly better at full resolution for the Small model (24.7\% to 22.6\%)
and worse for the Large model (20.5\% to 28.1\%). The cause is the metric scale (\cref{tab:scale}).
Depth Anything V2 places the scenes 1.45--1.54 times too close at its native input and 1.8--2.4 times
too close at full resolution, and Depth Anything~3 is almost unbiased at its native input (scale 0.98)
but places the scenes 1.5 times too far at full resolution. The models are trained at 0.2--0.7\,MP; at
full resolution each 14\,px patch covers much less of the scene, the token count is 8--32 times larger
and the positional embeddings are interpolated far beyond the training range, and nothing in the
networks ties the metric scale to a quantity that becomes more precise with more pixels.

With the scale removed by an oracle (\cref{fig:scaling}b) the shape does not improve either: the
oracle-scaled AbsRel stays flat or rises, from 6.0\% to 6.2\% for Depth Anything~3, from 3.8\% to
4.0\% for MoGe-2-L, from 8.3--8.9\% to 9.7--10.7\% for Depth Anything V2, from 3.9--5.7\% to 6.5--9.0\%
for UniDepth V2 and from 4.0--6.5\% to 5.2--8.4\% for Metric3D v2. Higher resolution sharpens
the monocular depth maps (\cref{sec:qualitative}), but their error is dominated by smooth deviations in
scale, tilt and curvature that come from the network's global interpretation of the scene, not from
missing image detail.

\subsection{Accuracy}
\label{sec:accuracy}
\Cref{tab:main} in \cref{app:results} lists accuracy, GPU time and energy for every configuration,
averaged over the 10 scenes.

\paragraph{Stereo accuracy.} Fast-FoundationStereo is the most accurate method: 0.33\% AbsRel with
8 iterations and 0.40\% with 4 at $768\times1152$, at 100\% density and 99\% of pixels within 5\% of
the true depth. It degrades gracefully with input size: each step down to $512\times768$ and
$384\times576$ cuts its GPU time by 2--3 times and adds only about 0.1 percentage points of AbsRel
(0.51\% and 0.64\% at $384\times576$). libSGM reaches 0.63\% AbsRel on 76\% of the non-occluded pixels
at full resolution, 0.73\% on 85\% at half and 1.18\% on 88\% at quarter resolution; each halving of
the image side cuts its GPU time by about four. At full resolution libSGM is about as accurate as
Fast-FoundationStereo at $384\times576$ (0.64\%) on the pixels it returns, at 1.5 times the GPU time
but without its density. After hole filling, libSGM reaches 1.92\%, 1.79\% and 2.13\% at full, half
and quarter resolution. The tuned smoothness penalties (\cref{app:libsgm}) lower libSGM's AbsRel by
39\%, 30\% and 9\% relative to its defaults, at 4--8 percentage points lower density. NVIDIA's ESS is
dense and reaches 0.70\% AbsRel at its fixed $576\times960$ input, and Light ESS 1.16\% at
$288\times480$, the best accuracy per unit of compute of all methods. Their bad-2 rates (20\% and
37\%) are higher than those of Fast-FoundationStereo (7--17\%) because a disparity error of one pixel at
their input is three or six pixels at full resolution. The commercially licensed
C-Fast-FoundationStereo matches the research checkpoint \texttt{23-36-37}, whose architecture it shares,
to within 0.01 percentage points of AbsRel at every input size (0.33\%, 0.43\% and 0.52\%), with a
slightly lower bad-2 rate at $768\times1152$ (6.0\% against 6.5\%) and the same GPU time.

\paragraph{Monocular accuracy.} No monocular model is close to metric accuracy at Middlebury's
1.3--7.6\,m depth range. The best raw results come from Depth Pro with the true focal length, 12.0\%
AbsRel (19.0\% with its own focal-length estimate), and from Depth Anything~3 Metric-Large, 12.3\% at a
fraction of the compute; only 29\% and 31\% of their pixels are within 5\% of the true depth. MoGe-2
reaches 19--33\% with the true focal length and 30--40\% with its own: its metric scale is unreliable
on these scenes, and with the true focal length it places parts of some scenes beyond the
zero-disparity distance (76--97\% density). Depth Anything V2 has 30--32\% AbsRel at its native input
and 41--56\% at full resolution, where only 0.1--2\% of its pixels are within 5\% of the true depth.
None of the monocular models benefits from a larger input: Depth Anything~3 has 14.2\% AbsRel at
$700\times1008$ and 54.9\% at full resolution, where it places 44\% of the pixels beyond the
zero-disparity distance, and MoGe-2 with the true focal length has 23--43\% at $1400\times2030$ and
24--40\% at full resolution, while its oracle-scaled error rises by 0.2--1.1 percentage points. All
numbers below are at native input unless stated otherwise. Removing the global scale error with an
oracle helps all of them, most of all MoGe-2, whose point maps have the best shape: with oracle scale
MoGe-2-L reaches 3.8\%, Depth Pro 4.7\%, Depth Anything~3 6.0\% and Depth Anything V2 8.3--8.9\%. Even
the best of these keeps 27\% of its pixels more than 5\% from the true depth. Of the two further
families, UniDepth V2, given the true intrinsics, has 20--26\% AbsRel at its native input (3.9--5.7\%
with oracle scale), and Metric3D v2, given the true focal length, 57--60\% (4.0--6.5\%): it places
most scenes too far and 35--37\% of the pixels beyond the zero-disparity distance.

\paragraph{Stereo versus monocular.} Every stereo method is more accurate than every monocular method,
raw or hole-filled, even when the monocular models are given an oracle scale: the worst stereo result
(hole-filled libSGM at quarter resolution, 2.13\%) is below the best oracle-scaled monocular result
(MoGe-2-L, 3.78\%). Without the oracle, the most accurate monocular model per unit of compute, Depth
Anything~3, has 10 times the AbsRel of libSGM at quarter resolution, which needs a fifth of its GPU
time, and 18 times that of ESS, which needs a quarter.

\subsection{Depth discontinuities}
\label{sec:edges}

\begin{figure*}[t]
\centering
\includegraphics[width=\linewidth]{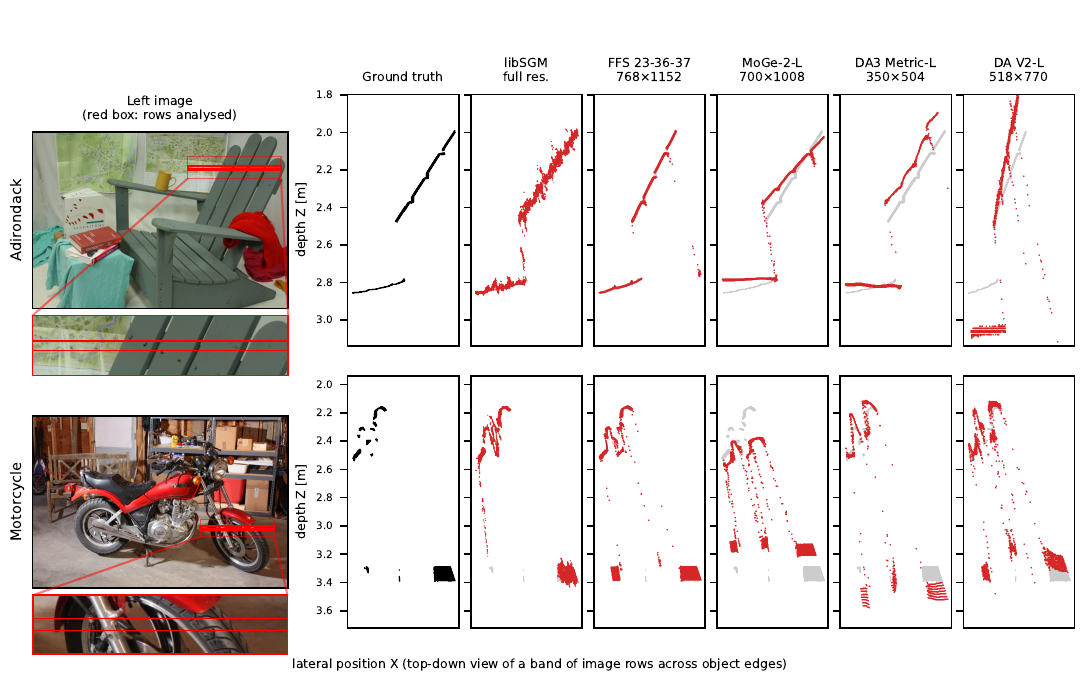}
\caption{Point clouds at object edges, seen from above: depth $Z$ against lateral position $X$ of the
points in a band of 40 image rows across the back slats of the Adirondack chair (top) and the front
wheel of the motorcycle (bottom); the left column shows the rectified left image with the band in red and, below it, a magnified view of
the band with some rows of context.
Grey: ground truth. Each method is shown at its own output resolution
(one point per output pixel); monocular depth with oracle scale. The monocular models connect
foreground and background with ramps of ``flying'' points; Fast-FoundationStereo keeps dense, clean
surfaces with a few flying points; libSGM at full resolution has the fewest flying points but noisier
surfaces and missing points at the edges.}
\label{fig:edges}
\end{figure*}

\begin{table}[t]
\centering
\caption{Depth discontinuities, mean over the 10 scenes. Edge pixels: non-occluded ground-truth pixels
whose $17\times17$ neighbourhood spans a depth ratio of at least 1.2 (1.7\% of the scored pixels).
Flying points: predicted edge pixels more than 5\% in front of the farthest and behind the nearest
ground-truth surface in that neighbourhood. Each method at its own output resolution (nearest-neighbour
upsampling); monocular depth with oracle scale.}
\label{tab:edges}
\small
\resizebox{\columnwidth}{!}{\begin{tabular}{@{}lrrrr@{}}
\toprule
Method (output resolution) & Density at edges [\%] & Flying points [\%] & AbsRel at edges [\%] & AbsRel elsewhere [\%] \\
\midrule
libSGM, full res. & 45 & 1.1 & 4.33 & 0.59 \\
libSGM, quarter res. & 59 & 9.4 & 9.08 & 1.08 \\
FFS 23-36-37, 768$\times$1152 & 100 & 3.0 & 2.48 & 0.29 \\
FFS 20-30-48, 768$\times$1152 & 100 & 4.1 & 2.85 & 0.36 \\
C-FFS, 768$\times$1152 & 100 & 2.4 & 2.42 & 0.29 \\
Depth Anything V2-S, 518$\times$770 & 100 & 25.8 & 11.43 & 8.90 \\
Depth Anything V2-L, 518$\times$770 & 100 & 20.3 & 11.94 & 8.26 \\
Depth Anything 3 Metric-L, 350$\times$504 & 100 & 16.6 & 10.25 & 5.92 \\
MoGe-2-L, 700$\times$1008 & 100 & 46.1 & 7.46 & 3.71 \\
ESS, 576$\times$960 & 100 & 3.8 & 4.51 & 0.63 \\
Light ESS, 288$\times$480 & 100 & 8.8 & 7.60 & 1.03 \\
Metric3D v2-L, 616$\times$1064 & 100 & 22.8 & 8.53 & 3.93 \\
UniDepth V2-L, 644$\times$938 & 100 & 17.1 & 7.83 & 4.03 \\
\bottomrule
\end{tabular}
}
\end{table}

AbsRel averages the error over all non-occluded pixels and is therefore dominated by large surfaces.
Depth discontinuities, where obstacle detection, grasping and free-space estimation need clean object
boundaries, make up only 1.7\% of the scored pixels. We analyse them separately (\cref{tab:edges},
\cref{fig:edges}). Each method is evaluated at its own output resolution, upsampled with
nearest-neighbour interpolation so that no intermediate depths come from our resampling, and
monocular depth with its oracle scale, so that a global scale error is not counted as an edge error.

The monocular models smooth across discontinuities and connect foreground and background with
continuous ramps of points (\cref{fig:edges}): 46\% of MoGe-2-L's edge points are flying points, 20--26\%
of Depth Anything V2's, 23\% of Metric3D v2-L's and 17\% of Depth Anything~3's and UniDepth V2-L's. Fast-FoundationStereo returns a point for every
pixel and keeps clean surfaces, with 2.4--4.1\% flying points (2.4\% for C-Fast-FoundationStereo),
mostly at thin structures such as the spokes of the wheel. libSGM at full resolution has the fewest flying points of all methods, 1.1\%: it
does not interpolate across an edge but rejects ambiguous pixels in its left--right check, so it returns
only 45\% of the edge pixels, and its surfaces are noisier. At quarter resolution libSGM has 9.4\% flying
points, more than Fast-FoundationStereo. ESS has 3.8\%, similar to Fast-FoundationStereo, and Light
ESS, at a quarter of ESS's input pixels, 8.8\%, similar to libSGM at quarter resolution.

AbsRel registers almost none of this. MoGe-2-L's AbsRel is 7.5\% at edge pixels and 3.7\% elsewhere,
but with 1.7\% of the pixels at edges, its flying points add less than 0.1 percentage points to its
overall error, and the oracle-scaled error that ranks MoGe-2 as the best monocular shape is the same
error that hides its edge behaviour. A dense, accurate-on-average depth map can therefore still
misplace object boundaries, and a classical matcher can be the better choice where boundaries matter:
at full resolution, libSGM's discontinuities are cleaner than those of the learned stereo network,
at the cost of missing pixels, a trade-off that a mean error over predicted pixels does not show.

\subsection{Qualitative results}
\label{sec:qualitative}
\begin{figure*}[t]
\centering
\includegraphics[width=\linewidth]{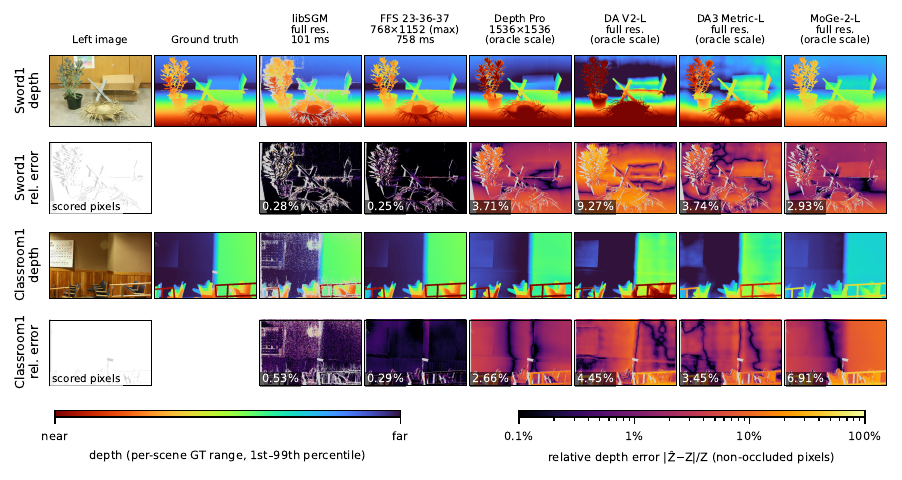}
\caption{Depth maps (rows 1 and 3) and relative depth error on non-occluded pixels (rows 2 and 4) for
Sword1 and Classroom1, with every method at full resolution or at its maximum or fixed input size.
Stereo: raw outputs. FFS: Fast-FoundationStereo; DA: Depth Anything. Monocular: depth after per-image oracle median scaling, so the remaining error is
structural, not a global scale error. Sword1 is the scene on which every method's error is closest to
its median over the 10 scenes; Classroom1 is added for its large planar surfaces. Numbers are the
AbsRel of each panel. Depth uses one colour scale per scene (1st--99th percentile of the ground truth);
values outside it are clipped. Grey: no ground truth, occluded, or no prediction (stereo holes are not
filled). The first column of the error rows shows the scored (non-occluded) pixels in white.}
\label{fig:depthmaps}
\end{figure*}

\Cref{fig:depthmaps} shows depth and error maps on two scenes for every
method at full resolution, or at its maximum or fixed input size, with the monocular networks after
oracle scaling, i.e.\ with their global scale error removed. Stereo errors are mostly below 1\% and
concentrate at depth discontinuities and thin structures (leaves, straw); libSGM leaves holes there and
on low-texture walls, and its full-resolution output is speckled where the left--right check rejects
isolated pixels. The oracle-scaled monocular depth maps look plausible and correctly ordered, yet keep
2.7--9.3\% error that is spatially structured: smooth bands across the walls and floor of Classroom1
that correspond to tilted or warped surfaces. This is the failure mode that a precision metric would
not see (\cref{sec:setup}). Without the oracle scale (\cref{tab:main}), the monocular errors are far
larger.

\begin{figure*}[t]
\centering
\includegraphics[width=\linewidth]{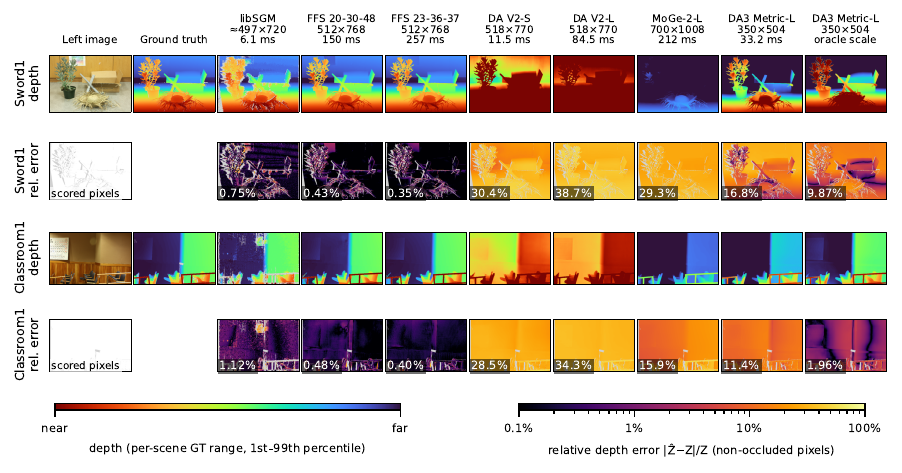}
\caption{As \cref{fig:depthmaps}, at low resolution: libSGM at quarter resolution, Fast-FoundationStereo
at $512\times768$ and the monocular models at their native input, each as deployed, i.e.\ monocular depth
with its own metric scale (last column: Depth Anything~3 with oracle scale for comparison). Column
titles give the processing resolution and GPU time per frame on Jetson AGX Orin.}
\label{fig:depthmaps_lowres}
\end{figure*}

\Cref{fig:depthmaps_lowres} compares the methods near the monocular models' native input, as deployed,
i.e.\ without oracle scale. libSGM at quarter resolution (6.1\,ms) keeps about 1\% error with holes at
occlusions and in low-texture regions; Fast-FoundationStereo at $512\times768$ is dense and has
0.35--0.48\% error. The monocular models at their native input have 11--39\% error on these scenes:
Depth Anything V2 places almost all of Sword1 at the near end of the depth range and MoGe-2-L at the
far end, and the error of Depth Anything~3 falls to 2.0--9.9\% only with the oracle scale.

\subsection{FP16 fidelity}
\label{sec:fp16}
We compared every monocular TensorRT FP16 engine with its FP32 PyTorch model on one scene. Depth
Anything V2 at native resolution differs by 0.10--0.17\%, and Base and Large at full resolution by
0.19\% and 0.25\% (median relative depth
difference). Depth Anything V2-Small at full resolution differs by 17.5\%, and Depth Pro by 1.9\% (with a
1.6$^\circ$ shift of its field-of-view estimate). For these two we report accuracy from FP32 and speed
from FP16, which favours them. The FP32 Small model is itself inaccurate at this resolution (57.7\%
AbsRel), so the FP16 error is not the cause of its poor result.

For Depth Anything~3 and MoGe-2 we found two TensorRT 10.3 problems, both in fused kernels: with
every intermediate tensor marked as an output the engines are correct, and ONNX Runtime reproduces
PyTorch to within 0.002\%. First, the attention pattern that PyTorch exports from
\texttt{scaled\_dot\_product\_attention} (queries and keys each scaled by the square root of the scale
factor) produced a wrong engine; exporting the same computation as
$\operatorname{softmax}(q\,s\,k^{\top})\,v$ fixed it. The Depth Anything V2 full-resolution
engines use the same attention export; Base and Large match FP32, but the 17.5\% deviation of the
Small engine is consistent with this problem, so its timing may not represent a correct engine. Second, MoGe-2-S and -B's three-layer metric-scale
head on the class token came out wrong even in FP32; we run this negligible computation outside the
engine. After these fixes the Depth Anything~3 engine reproduces the FP32 AbsRel to within
0.2 percentage points on the check scenes processed at its $350\times504$ input size, and the MoGe-2
engines differ from FP32 by 0.03--1.1\% (median relative depth). For the larger inputs, up to
${\approx}$29{,}000 tokens at full resolution, the export processes attention in blocks of 2{,}048
queries, which is the same computation without tracing the full attention matrix. Their accuracy is reported from FP32 and their speed from the fixed FP16 engines.

The UniDepth V2 engines, exported with the same attention form, agree with FP32 to within 0.05--0.49\%
(median relative depth difference). The Metric3D v2 engines differ by 1.3--3.3\% (99th percentile up
to 15\%), probably because its iterative refinement decoder accumulates FP16 rounding; for both we
report accuracy from FP32 and speed from FP16.

\section{Accuracy per compute}
\label{sec:compute}
\begin{figure*}[t]
\centering
\includegraphics[width=\linewidth]{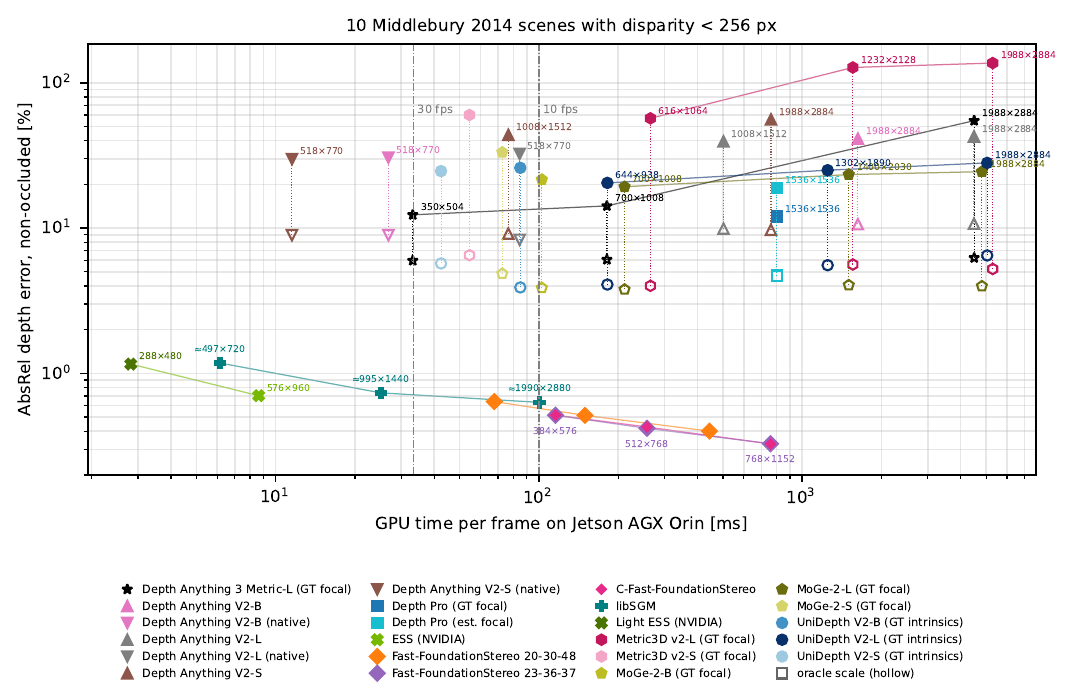}
\caption{Depth error (AbsRel on non-occluded pixels, log scale) against GPU time per frame on Jetson
AGX Orin, averaged over the 10 scenes. Filled
markers are the methods as deployed (raw output; hole-filled results are in \cref{tab:main} only);
hollow markers are monocular methods with an oracle per-image scale. Upward triangles are Depth
Anything V2 at full resolution, downward triangles at its native $518\times770$ input. The black stars are Depth Anything~3 Metric-L and the olive
pentagons MoGe-2-S/B/L, the pink hexagons Metric3D v2-S/L and the blue circles UniDepth V2-S/B/L,
with the ground-truth focal length or intrinsics; thin lines connect the native, twice-native and
full-resolution runs of Depth Anything~3, MoGe-2-L, Metric3D v2-L and UniDepth V2-L (the smaller
models at other sizes and MoGe-2 with its own focal estimate are in \cref{tab:main} only). Green
crosses are ESS and Light ESS, connected by a thin line; thin lines also connect the three input
resolutions of libSGM (teal crosses) and of each Fast-FoundationStereo model. The small pink diamonds
(C-Fast-FoundationStereo) coincide with the purple ones of the research checkpoint \texttt{23-36-37}.
Labels give the
processing resolution (height$\times$width). Vertical lines mark the per-frame budgets for 30 and 10\,fps on one
Orin.}
\label{fig:absrel}
\end{figure*}

\Cref{fig:absrel} plots depth error against GPU time for every configuration (raw outputs only).

\paragraph{Speed.} GPU time spans more than three orders of magnitude, from 2.8\,ms to 5.3\,s per
frame. Among the stereo methods, Light ESS takes 2.8\,ms and ESS 8.6\,ms; libSGM 6.1\,ms at quarter,
25\,ms at half and 101\,ms at full resolution; and Fast-FoundationStereo 68\,ms (4 iterations) and
116\,ms (8 iterations) at $384\times576$, rising to 445\,ms and 758\,ms at $768\times1152$
(C-Fast-FoundationStereo takes the same time as the 8-iteration research checkpoint). Among the
monocular models at their native input, Depth Anything V2 takes 11.5\,ms (Small), 26.8\,ms (Base) and
84.5\,ms (Large), Depth Anything~3 33\,ms, UniDepth V2 42\,ms (Small), 85\,ms (Base) and 182\,ms
(Large), Metric3D v2 54\,ms (Small) and 265\,ms (Large) and MoGe-2 73\,ms, 103\,ms and 212\,ms. At twice
that size, Depth Anything~3 takes 181\,ms, UniDepth V2 0.28--1.25\,s, Metric3D v2 0.30--1.56\,s and
MoGe-2 0.42--1.50\,s. At full or fixed resolution, Depth Pro takes 803\,ms and the other monocular
models 0.76--5.3\,s. No monocular model is faster than Light ESS, ESS or libSGM at quarter resolution:
the fastest, Depth Anything V2-S at native input, needs 4.1 times the GPU time of Light ESS, Depth
Anything~3 12 times, and the monocular models at full or fixed resolution 272--1{,}890 times.

\paragraph{Energy.} Dynamic energy per frame is 0.08\,J for Light ESS, 0.31\,J for ESS, 0.18, 0.73 and
3.0\,J for libSGM at quarter, half and full resolution, and 1.9--22.4\,J for Fast-FoundationStereo. The
monocular models need 0.5--12\,J at their native input and 32--235\,J at full or fixed resolution. The
networks draw 34--57\,W on the GPU+SoC and CPU rails under load.

\subsection{Cost of compute versus a second camera}
\label{sec:cost}
Running a method at $F$ frames/s uses a fraction $u=t_\text{GPU}F/1000$ of one Orin, where
$t_\text{GPU}$ is the per-frame GPU time in milliseconds. We price this at the module's volume price
$P_\text{SoC}$ (\$2{,}999 for the AGX Orin 64GB at 1k units~\cite{nvidia2026jetsonprice}). A stereo
system with GPU time $t_s$ is cheaper than a monocular system with GPU time $t_m$ whenever the extra
camera costs less than
\begin{equation}
P^{*}_\text{cam} = (t_m - t_s)\,\frac{F}{1000}\,P_\text{SoC}.
\label{eq:breakeven}
\end{equation}
We compare $P^{*}_\text{cam}$ with camera prices of \$20 to \$500 per camera; for reference, automotive
camera modules average roughly \$20 each across the market~\cite{yole2025imaging}. \Cref{tab:cost}
applies the model at 10 and 30\,fps.

\begin{table*}[t]
\centering
\caption{System cost: cameras plus the fraction of Jetson AGX Orin 64GB GPU time used (\$2{,}999 per
module). Stereo uses two cameras and monocular one. ``Orins'' above 1 require multiple modules (or a
larger GPU) for that frame rate. Break-even: the camera price below which adding the second camera
and running Light ESS, the cheapest stereo configuration, is cheaper than the
monocular model (\cref{eq:breakeven}). Monocular models with their own focal-length estimate have the
same cost as with the true one and are omitted.}
\label{tab:cost}
\small
\resizebox{\textwidth}{!}{\begin{tabular}{llrrrrr}
\toprule
Method & Resolution & Orins @10\,fps & Orins @30\,fps & \${}@30\,fps, \$20 per camera & \${}@30\,fps, \$500 per camera & Break-even camera @30\,fps \\
\midrule
Light ESS (NVIDIA) & 288$\times$480 (fixed) & 0.03 & 0.08 & 293 & 1,253 & -- \\
libSGM & $\approx$497$\times$720 (quarter) & 0.06 & 0.18 & 592 & 1,552 & -- \\
ESS (NVIDIA) & 576$\times$960 (fixed) & 0.09 & 0.26 & 816 & 1,776 & -- \\
libSGM & $\approx$995$\times$1440 (half) & 0.25 & 0.75 & 2,297 & 3,257 & -- \\
Fast-FoundationStereo 20-30-48 & 384$\times$576 & 0.68 & 2.03 & 6,132 & 7,092 & -- \\
libSGM & full ($\approx$1990$\times$2880) & 1.01 & 3.02 & 9,090 & 10,050 & -- \\
Fast-FoundationStereo 23-36-37 & 384$\times$576 & 1.16 & 3.47 & 10,445 & 11,405 & -- \\
C-Fast-FoundationStereo & 384$\times$576 & 1.16 & 3.47 & 10,446 & 11,406 & -- \\
Fast-FoundationStereo 20-30-48 & 512$\times$768 & 1.50 & 4.49 & 13,505 & 14,465 & -- \\
Fast-FoundationStereo 23-36-37 & 512$\times$768 & 2.57 & 7.72 & 23,204 & 24,164 & -- \\
C-Fast-FoundationStereo & 512$\times$768 & 2.58 & 7.73 & 23,218 & 24,178 & -- \\
Fast-FoundationStereo 20-30-48 & 768$\times$1152 (max) & 4.45 & 13.34 & 40,054 & 41,014 & -- \\
Fast-FoundationStereo 23-36-37 & 768$\times$1152 (max) & 7.58 & 22.74 & 68,247 & 69,207 & -- \\
C-Fast-FoundationStereo & 768$\times$1152 (max) & 7.60 & 22.81 & 68,454 & 69,414 & -- \\
Depth Anything V2-S (native) & 518$\times$770 (native) & 0.12 & 0.35 & 1,058 & 1,538 & \$784 \\
Depth Anything V2-B (native) & 518$\times$770 (native) & 0.27 & 0.80 & 2,433 & 2,913 & \$2,159 \\
Depth Anything 3 Metric-L (GT focal) & 350$\times$504 (native) & 0.33 & 1.00 & 3,005 & 3,485 & \$2,732 \\
UniDepth V2-S (GT intrinsics) & 644$\times$938 (native) & 0.43 & 1.28 & 3,844 & 4,324 & \$3,571 \\
Metric3D v2-S (GT focal) & 616$\times$1064 (native) & 0.54 & 1.63 & 4,922 & 5,402 & \$4,649 \\
MoGe-2-S (GT focal) & 700$\times$1008 (native) & 0.73 & 2.18 & 6,553 & 7,033 & \$6,279 \\
Depth Anything V2-S & 1008$\times$1512 & 0.77 & 2.30 & 6,908 & 7,388 & \$6,635 \\
Depth Anything V2-L (native) & 518$\times$770 (native) & 0.84 & 2.53 & 7,621 & 8,101 & \$7,348 \\
UniDepth V2-B (GT intrinsics) & 644$\times$938 (native) & 0.85 & 2.55 & 7,669 & 8,149 & \$7,395 \\
MoGe-2-B (GT focal) & 700$\times$1008 (native) & 1.03 & 3.08 & 9,253 & 9,733 & \$8,980 \\
Depth Anything 3 Metric-L (GT focal) & 700$\times$1008 & 1.81 & 5.44 & 16,330 & 16,810 & \$16,056 \\
UniDepth V2-L (GT intrinsics) & 644$\times$938 (native) & 1.82 & 5.46 & 16,400 & 16,880 & \$16,126 \\
MoGe-2-L (GT focal) & 700$\times$1008 (native) & 2.12 & 6.35 & 19,056 & 19,536 & \$18,783 \\
Metric3D v2-L (GT focal) & 616$\times$1064 (native) & 2.65 & 7.96 & 23,890 & 24,370 & \$23,617 \\
UniDepth V2-S (GT intrinsics) & 1302$\times$1890 & 2.79 & 8.37 & 25,133 & 25,613 & \$24,860 \\
Metric3D v2-S (GT focal) & 1232$\times$2128 & 2.96 & 8.87 & 26,612 & 27,092 & \$26,339 \\
MoGe-2-S (GT focal) & 1400$\times$2030 & 4.19 & 12.56 & 37,695 & 38,175 & \$37,421 \\
Depth Anything V2-L & 1008$\times$1512 & 5.03 & 15.08 & 45,251 & 45,731 & \$44,977 \\
UniDepth V2-B (GT intrinsics) & 1302$\times$1890 & 5.80 & 17.39 & 52,169 & 52,649 & \$51,896 \\
MoGe-2-B (GT focal) & 1400$\times$2030 & 6.63 & 19.90 & 59,710 & 60,190 & \$59,437 \\
Depth Anything V2-S & 1988$\times$2884 (full) & 7.62 & 22.85 & 68,539 & 69,019 & \$68,265 \\
Depth Pro (GT focal) & 1536$\times$1536 (fixed) & 8.03 & 24.08 & 72,221 & 72,701 & \$71,948 \\
Metric3D v2-S (GT focal) & 1988$\times$2884 (full) & 9.92 & 29.77 & 89,301 & 89,781 & \$89,028 \\
UniDepth V2-S (GT intrinsics) & 1988$\times$2884 (full) & 11.33 & 34.00 & 101,987 & 102,467 & \$101,713 \\
MoGe-2-S (GT focal) & 1988$\times$2884 (full) & 11.79 & 35.36 & 106,062 & 106,542 & \$105,789 \\
UniDepth V2-L (GT intrinsics) & 1302$\times$1890 & 12.50 & 37.50 & 112,492 & 112,972 & \$112,219 \\
MoGe-2-L (GT focal) & 1400$\times$2030 & 15.03 & 45.10 & 135,274 & 135,754 & \$135,000 \\
Metric3D v2-L (GT focal) & 1232$\times$2128 & 15.59 & 46.76 & 140,244 & 140,724 & \$139,970 \\
Depth Anything V2-B & 1988$\times$2884 (full) & 16.28 & 48.85 & 146,528 & 147,008 & \$146,255 \\
MoGe-2-B (GT focal) & 1988$\times$2884 (full) & 20.07 & 60.21 & 180,591 & 181,071 & \$180,317 \\
UniDepth V2-B (GT intrinsics) & 1988$\times$2884 (full) & 23.35 & 70.06 & 210,120 & 210,600 & \$209,846 \\
Depth Anything V2-L & 1988$\times$2884 (full) & 44.97 & 134.91 & 404,605 & 405,085 & \$404,332 \\
Depth Anything 3 Metric-L (GT focal) & 1988$\times$2884 (full) & 45.04 & 135.11 & 405,218 & 405,698 & \$404,944 \\
MoGe-2-L (GT focal) & 1988$\times$2884 (full) & 48.18 & 144.54 & 433,486 & 433,966 & \$433,213 \\
UniDepth V2-L (GT intrinsics) & 1988$\times$2884 (full) & 50.48 & 151.43 & 454,144 & 454,624 & \$453,870 \\
Metric3D v2-L (GT focal) & 1988$\times$2884 (full) & 52.98 & 158.93 & 476,638 & 477,118 & \$476,365 \\
\bottomrule
\end{tabular}
}
\end{table*}

The cheapest stereo configuration, Light ESS, needs 0.08 Orins at 30\,fps, ESS 0.26 and libSGM
0.18--3.0; Fast-FoundationStereo needs 2.0--22.7. The monocular models need 0.35 (Depth Anything V2-S)
to 8.0 (Metric3D v2-L) Orins at their native input and 22.8 to 159 at full or fixed resolution.

Relative to Light ESS, the break-even camera price at 30\,fps is \$786 (Depth Anything V2-S) to
\$23{,}618 (Metric3D v2-L) for the monocular models at their native input and \$68{,}266 to \$476{,}366 at
full or fixed resolution (\$71{,}949 for Depth Pro). Every one of them is above the whole \$20--\$500
range of camera prices, so at 30\,fps a second camera plus Light ESS is the cheaper system even with a
\$500 camera. At 10\,fps the thresholds are a third as large; they remain above \$500 for every monocular
model except Depth Anything V2-S at native input, whose threshold of \$262 falls within the range. Against ESS, the full- or fixed-resolution thresholds at 30\,fps
are \$67{,}748 to \$475{,}845. Fast-FoundationStereo, the most accurate method, needs more GPU time than
several monocular models at native input but has 19--182 times lower AbsRel: learned stereo buys
accuracy with compute, while an efficient stereo matcher is what makes the second camera economical.

\section{Discussion and limitations}

\paragraph{Why stereo scales and monocular depth does not.} Stereo measures depth: the disparity of
a point is observed in the images, and the calibrated focal length and baseline turn it into metric
depth, with an error that shrinks as the pixels become smaller. A monocular network infers depth from
learned regularities of the scene, and its metric scale comes from a prior over object sizes and camera
geometry seen in training. More pixels add detail but no information about that scale, and outside the
training resolution the prior itself shifts (\cref{tab:scale}). The two approaches therefore move apart
as cameras gain resolution.

\paragraph{Running monocular models beyond their training resolution.} One may object that the
monocular models were never meant to run at 5.7\,MP. That is the point of the comparison: stereo turns
additional pixels into metric accuracy, and the released monocular models cannot. They are also far
less accurate at their native input, where the best of them, Depth Anything~3, has 18 times the error
of ESS, which needs a quarter of its GPU time, and still 8.5 times that error with an oracle scale. Depth Pro, whose multi-scale design targets high resolution, reaches 12.0\% at its fixed
$1536\times1536$ input. Tile-based high-resolution methods~\cite{li2024patchfusion,li2024patchrefiner}
sharpen detail by fusing crops with a low-resolution global estimate from which they inherit the metric
scale; we expect them to sharpen edges rather than remove the scale error, but we have not evaluated
them.

\paragraph{Licences.} Licences differ more than accuracy might suggest (\cref{tab:res}). libSGM
(Apache-2.0), Depth Anything V2-S and Depth Anything~3 Metric-Large (Apache-2.0), MoGe-2 (MIT) and
Metric3D v2 (BSD-2-Clause) are permissive; ESS is distributed under NVIDIA's model EULA, whose model
card states that it is ready for commercial use; C-Fast-FoundationStereo is governed by the NVIDIA Open
Model Agreement, although its model card also describes it as intended for research and evaluation;
the Fast-FoundationStereo research checkpoints (NVIDIA Source Code License),
Depth Anything V2-B/L and UniDepth V2 (CC BY-NC 4.0) are licensed for non-commercial use only, and
Depth Pro under Apple's own licence. We report the licences as the authors state them; they are not
legal advice.

\paragraph{Coverage.} The benchmark covers the model families we consider most relevant for metric depth
on an Orin as of October 2026, not every published method. We did not evaluate the Orin's optical flow
accelerator through VPI, FoundationStereo itself (Fast-FoundationStereo is its real-time version),
other learned stereo networks such as RAFT-Stereo, or tile-based high-resolution monocular methods.

\paragraph{Edges and the limits of AbsRel.} The mean AbsRel over predicted pixels rewards dense,
smooth output and is nearly blind to depth discontinuities (\cref{sec:edges}). It also treats a missing
pixel as neutral, which favours sparse stereo, and a flying point as a small error, which favours the
networks. Neither effect changes the ranking between stereo and monocular methods here, but it means
that the most accurate method by AbsRel is not necessarily the best at object boundaries: a classical
matcher at full resolution produces the cleanest boundaries of all methods we tested.

\paragraph{Scope of the data.} Our 10 Middlebury 2014 scenes are indoor and close-range (1.3--7.6\,m),
with 17--24\,cm baselines and focal lengths of 2{,}800--7{,}200\,px. Disparities are therefore large and
stereo triangulation is very accurate. Stereo depth error grows quadratically with range, so at long
range with a short baseline the gap would shrink; wide-baseline automotive stereo is designed to
counter this. A long-range, outdoor benchmark with LiDAR ground truth is the most important follow-up.

\paragraph{Fairness to monocular methods.} We gave the monocular models the true focal length or
intrinsics (Depth Pro, Depth Anything~3, MoGe-2, Metric3D v2, UniDepth V2), a domain-matched indoor checkpoint (Depth Anything V2 Hypersim) and an
oracle scale, none of which is available in deployment. For MoGe-2 the true focal length helps the
shape (lower oracle-scaled error) but not the metric scale, which comes from a separate head. Evaluating on non-occluded pixels removes the 9\% of the ground truth that
stereo cannot triangulate by construction; the restriction favours stereo only by excluding those
pixels, which the hole-filled rows of \cref{tab:main} account for. We did not fine-tune any model on Middlebury. We cannot rule out that Middlebury images appear in any model's training data; if they do,
that would favour the networks. A model fine-tuned for a specific camera would close part of the
raw-to-oracle gap but not the oracle gap itself.

\paragraph{Disparity range and scene selection.} libSGM searches at most 256 disparity levels at its
processing resolution, which at full resolution covers 255\,px. We therefore restrict the evaluation to
the 10 scenes whose non-occluded disparity stays below that range; in the other 13 public scenes, near
objects at 265--712\,px of disparity would be out of range for libSGM at full resolution, which would
turn a search-range setting into a large accuracy penalty. This selection favours stereo only in the
sense that it removes a configuration problem with a known fix: running the matcher at lower
resolution (at half resolution 256 levels span ${\approx}$510\,px, at quarter resolution
${\approx}$1020\,px), or a matcher with a larger search range. It also removes the closest-range
scenes, in which stereo depth is most accurate.

\paragraph{Parameter tuning.} libSGM's $P_1$ and $P_2$ were tuned on the evaluation scenes, which
favours it; cross-validation with one scene held out raises its AbsRel by only 0.01--0.08 percentage
points (\cref{app:libsgm}), so the conclusions do not depend on it. The networks use their released
weights without any tuning.

\paragraph{Timing methodology.} Network timings exclude image resizing, normalisation and
post-processing (for MoGe-2 the focal-length and shift recovery and the metric-scale head), while
the SGM timings include copying the images to and the disparity from the GPU, so the comparison
slightly favours the networks. The fractional cost model assumes compute can be shared across tasks; with dedicated
hardware, cost rises in steps of whole modules.

\section{Conclusion}
On a Jetson AGX Orin, stereo gives metric depth that no current monocular foundation model approaches.
At 30\,fps, NVIDIA's ESS reaches 0.70\% AbsRel with dense output in 8.6\,ms and libSGM 0.73\% in
25\,ms; with more compute Fast-FoundationStereo reaches 0.33\%, as does its commercially licensed
checkpoint. The best
monocular models, given the true focal length, reach 12\% (Depth Anything~3 Metric-L in 33\,ms, Depth
Pro in 803\,ms), and only with an oracle per-image scale do MoGe-2-L, UniDepth V2-B and Metric3D v2-L
reach 3.8--4.0\%, still several times the error of stereo. Stereo error falls as the input resolution
grows, while no monocular model improves, because its metric scale drifts outside the training
resolution and its shape does not sharpen; and at object boundaries the monocular models scatter
17--46\% of the points between surfaces, an error that AbsRel barely registers and where a classical
matcher at full resolution is the cleanest of all. For metric depth on an Orin, a second camera with
ESS, libSGM or Fast-FoundationStereo is the better choice at every compute budget we measured;
monocular models are the choice when only one camera is possible and the metric scale can be recovered
by other means.

\section*{Conflict of interest}
The author is employed by NODAR Inc., which develops and sells stereo vision products, and has a
financial interest in stereo vision. For fairness, we did not include NODAR's own stereo matcher,
Hammerhead, in the comparison: no NODAR product is evaluated here, and every method is open source or
publicly released and is run with publicly documented settings on the same hardware and the same
public ground truth. All evaluation code, per-scene results and engine build scripts are released, and
the monocular methods are additionally given the true focal length and an oracle scale.

\section*{Acknowledgements}
We thank the authors of Middlebury 2014, Depth Anything V2 and 3, Depth Pro, MoGe-2, Metric3D v2,
UniDepth V2 and Fast-FoundationStereo, NVIDIA for ESS and C-Fast-FoundationStereo, and Fixstars for
libSGM, for releasing their
data, code and weights.

{\small
\bibliographystyle{ieeetr}
\bibliography{refs}

\begin{thebibliography}{10}

\bibitem{yang2024depthv2}
L.~Yang, B.~Kang, Z.~Huang, Z.~Zhao, X.~Xu, J.~Feng, and H.~Zhao, ``{Depth
  Anything V2},'' in {\em Advances in Neural Information Processing Systems
  (NeurIPS)}, 2024.

\bibitem{lin2026da3}
H.~Lin, S.~Chen, J.~H. Liew, D.~Y. Chen, Z.~Li, {\em et~al.}, ``{Depth Anything
  3}: Recovering the visual space from any views,'' in {\em International
  Conference on Learning Representations (ICLR)}, 2026.

\bibitem{bochkovskii2025depthpro}
A.~Bochkovskii, A.~Delaunoy, H.~Germain, M.~Santos, Y.~Zhou, S.~R. Richter, and
  V.~Koltun, ``{Depth Pro}: Sharp monocular metric depth in less than a
  second,'' in {\em International Conference on Learning Representations
  (ICLR)}, 2025.

\bibitem{wang2025moge2}
R.~Wang, S.~Xu, Y.~Dong, Y.~Deng, J.~Xiang, Z.~Lv, G.~Sun, X.~Tong, and
  J.~Yang, ``{MoGe-2}: Accurate monocular geometry with metric scale and sharp
  details,'' in {\em Advances in Neural Information Processing Systems
  (NeurIPS)}, 2025.

\bibitem{hu2024metric3dv2}
M.~Hu, W.~Yin, C.~Zhang, Z.~Cai, X.~Long, H.~Chen, K.~Wang, G.~Yu, C.~Shen, and
  S.~Shen, ``{Metric3D v2}: A versatile monocular geometric foundation model
  for zero-shot metric depth and surface normal estimation,'' {\em IEEE
  Transactions on Pattern Analysis and Machine Intelligence}, 2024.

\bibitem{piccinelli2025unidepthv2}
L.~Piccinelli, C.~Sakaridis, Y.-H. Yang, M.~Segu, S.~Li, W.~Abbeloos, and
  L.~Van~Gool, ``{UniDepthV2}: Universal monocular metric depth estimation made
  simpler,'' {\em arXiv preprint arXiv:2502.20110}, 2025.

\bibitem{hirschmuller2008sgm}
H.~Hirschm{\"u}ller, ``Stereo processing by semiglobal matching and mutual
  information,'' {\em IEEE Transactions on Pattern Analysis and Machine
  Intelligence}, vol.~30, no.~2, pp.~328--341, 2008.

\bibitem{nvidia2024ess}
{NVIDIA}, ``{ESS} ({E}fficient {S}upervised {S}tereo) {DNN} stereo disparity
  model, version 4.1.0.'' NGC model
  \texttt{nvidia/isaac/dnn\_stereo\_disparity},
  \url{https://catalog.ngc.nvidia.com/orgs/nvidia/teams/isaac/models/dnn_stereo_disparity},
  2024.

\bibitem{wen2025fastfoundationstereo}
B.~Wen, S.~Dewan, and S.~Birchfield, ``{Fast-FoundationStereo}: Real-time
  zero-shot stereo matching,'' {\em arXiv preprint arXiv:2512.11130}, 2025.

\bibitem{scharstein2014middlebury}
D.~Scharstein, H.~Hirschm{\"u}ller, Y.~Kitajima, G.~Krathwohl,
  N.~Ne{\v{s}}i{\'c}, X.~Wang, and P.~Westling, ``High-resolution stereo
  datasets with subpixel-accurate ground truth,'' in {\em German Conference on
  Pattern Recognition (GCPR)}, 2014.

\bibitem{ranftl2022midas}
R.~Ranftl, K.~Lasinger, D.~Hafner, K.~Schindler, and V.~Koltun, ``Towards
  robust monocular depth estimation: Mixing datasets for zero-shot
  cross-dataset transfer,'' {\em IEEE Transactions on Pattern Analysis and
  Machine Intelligence}, vol.~44, no.~3, pp.~1623--1637, 2022.

\bibitem{ranftl2021dpt}
R.~Ranftl, A.~Bochkovskiy, and V.~Koltun, ``Vision transformers for dense
  prediction,'' in {\em IEEE/CVF International Conference on Computer Vision
  (ICCV)}, 2021.

\bibitem{yang2024depth}
L.~Yang, B.~Kang, Z.~Huang, X.~Xu, J.~Feng, and H.~Zhao, ``{Depth Anything}:
  Unleashing the power of large-scale unlabeled data,'' in {\em IEEE/CVF
  Conference on Computer Vision and Pattern Recognition (CVPR)}, 2024.

\bibitem{oquab2024dinov2}
M.~Oquab, T.~Darcet, T.~Moutakanni, {\em et~al.}, ``{DINOv2}: Learning robust
  visual features without supervision,'' {\em Transactions on Machine Learning
  Research}, 2024.

\bibitem{roberts2021hypersim}
M.~Roberts, J.~Ramapuram, A.~Ranjan, A.~Kumar, M.~A. Bautista, N.~Paczan,
  R.~Webb, and J.~M. Susskind, ``{Hypersim}: A photorealistic synthetic dataset
  for holistic indoor scene understanding,'' in {\em IEEE/CVF International
  Conference on Computer Vision (ICCV)}, 2021.

\bibitem{li2024patchfusion}
Z.~Li, S.~F. Bhat, and P.~Wonka, ``{PatchFusion}: An end-to-end tile-based
  framework for high-resolution monocular metric depth estimation,'' in {\em
  IEEE/CVF Conference on Computer Vision and Pattern Recognition (CVPR)}, 2024.

\bibitem{li2024patchrefiner}
Z.~Li, S.~F. Bhat, and P.~Wonka, ``{PatchRefiner}: Leveraging synthetic data
  for real-domain high-resolution monocular metric depth estimation,'' in {\em
  European Conference on Computer Vision (ECCV)}, 2024.

\bibitem{fixstars2024libsgm}
{Fixstars Corporation}, ``{libSGM}: Stereo semi-global matching on {CUDA}.''
  \url{https://github.com/fixstars/libSGM}, 2024.
\newblock Version 3.1.0, commit e4c669b.

\bibitem{lipson2021raft}
L.~Lipson, Z.~Teed, and J.~Deng, ``{RAFT-Stereo}: Multilevel recurrent field
  transforms for stereo matching,'' in {\em International Conference on 3D
  Vision (3DV)}, 2021.

\bibitem{wen2025foundationstereo}
B.~Wen, M.~Trepte, J.~Aribido, J.~Kautz, O.~Gallo, and S.~Birchfield,
  ``{FoundationStereo}: Zero-shot stereo matching,'' in {\em IEEE/CVF
  Conference on Computer Vision and Pattern Recognition (CVPR)}, 2025.

\bibitem{wang2019pseudolidar}
Y.~Wang, W.-L. Chao, D.~Garg, B.~Hariharan, M.~Campbell, and K.~Q. Weinberger,
  ``{Pseudo-LiDAR} from visual depth estimation: Bridging the gap in {3D}
  object detection for autonomous driving,'' in {\em IEEE/CVF Conference on
  Computer Vision and Pattern Recognition (CVPR)}, 2019.

\bibitem{eigen2014depth}
D.~Eigen, C.~Puhrsch, and R.~Fergus, ``Depth map prediction from a single image
  using a multi-scale deep network,'' in {\em Advances in Neural Information
  Processing Systems (NIPS)}, 2014.

\bibitem{nvidia2026jetsonprice}
{CNX Software}, ``{NVIDIA} increases the price of {Jetson} modules and devkits
  by up to 101\%.''
  \url{https://www.cnx-software.com/2026/07/22/nvidia-increases-the-price-of-jetson-modules-and-devkits-by-up-to-101/},
  2026.
\newblock Accessed 2026-10-10.

\bibitem{yole2025imaging}
{Yole Group}, ``Automotive imaging market surges toward \$8.7{B} milestone.''
  \url{https://www.edge-ai-vision.com/2025/07/automotive-imaging-market-surges-toward-8-7b-milestone/},
  2025.
\newblock Accessed 2026-10-10.

\end{thebibliography}
}

\clearpage
\onecolumn
\appendix
\section{Dataset}
\label{app:dataset}
\Cref{tab:dataset} lists the 10 Middlebury 2014 scenes~\cite{scharstein2014middlebury} that we use: the
scenes with public ground truth whose non-occluded disparity stays below 256\,px, the range of the
256-level semi-global matcher at full resolution. We use them at full resolution with ``perfect'' rectification.
Ground-truth disparity comes from structured light and is given in full-resolution pixels;
\texttt{ndisp} is the search bound in Middlebury's calibration file, not the true maximum.
\emph{Non-occ.\ max} is the largest disparity among pixels visible in both views, which is the ground
truth we evaluate. \emph{Occl.} and \emph{Out} are the fractions of ground-truth pixels occluded in the
right view or outside its field of view. The other 13 public scenes (Cable, Couch, Flowers, Jadeplant,
Mask, Pipes, Playroom, Playtable, Shopvac, Sticks, Storage, Sword2, Vintage) have non-occluded
disparities of 265--712\,px and are not used.

\begin{table}[!h]
\centering
\caption{The 10 Middlebury 2014 scenes used (left image thumbnails). Disparities in
full-resolution pixels; depth from $Z=fB/(d+d_\text{offs})$.}
\label{tab:dataset}
\small
\resizebox{\textwidth}{!}{\begin{tabular}{@{}cl r r rrrr r r rr@{}}
\toprule
 & & & & \multicolumn{4}{c}{GT disparity [px]} & Non-occ. & Depth & Occl. & Out \\
\cmidrule(lr){5-8}
Image & Scene & Size ($W\times H$) & \texttt{ndisp} & min & median & 99\,\% & max & max [px] & [m] & [\%] & [\%] \\
\midrule
\raisebox{-0.45\height}{\includegraphics[height=0.72cm]{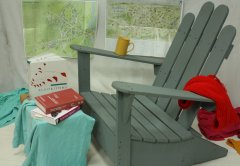}} & Adirondack & 2880$\times$1988 & 280 & 30 & 108 & 219 & 243 & 243 & 1.62--3.07 & 6.1 & 2.0 \\
\raisebox{-0.45\height}{\includegraphics[height=0.72cm]{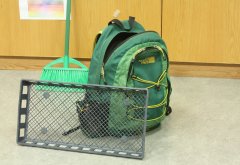}} & Backpack & 2940$\times$2016 & 260 & 32 & 130 & 214 & 224 & 224 & 2.22--3.36 & 8.4 & 3.1 \\
\raisebox{-0.45\height}{\includegraphics[height=0.72cm]{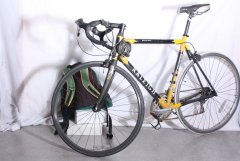}} & Bicycle1 & 2988$\times$2008 & 180 & 56 & 73 & 127 & 145 & 145 & 2.94--4.08 & 6.3 & 2.6 \\
\raisebox{-0.45\height}{\includegraphics[height=0.72cm]{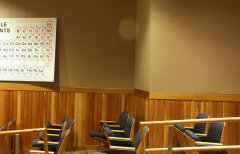}} & Classroom1 & 3000$\times$1920 & 250 & 16 & 55 & 203 & 208 & 208 & 2.98--7.63 & 2.7 & 1.8 \\
\raisebox{-0.45\height}{\includegraphics[height=0.72cm]{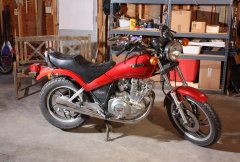}} & Motorcycle & 2964$\times$2000 & 270 & 28 & 152 & 231 & 240 & 240 & 2.11--5.04 & 7.9 & 3.1 \\
\raisebox{-0.45\height}{\includegraphics[height=0.72cm]{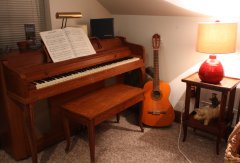}} & Piano & 2820$\times$1920 & 260 & 38 & 104 & 184 & 209 & 209 & 1.51--3.11 & 5.0 & 3.6 \\
\raisebox{-0.45\height}{\includegraphics[height=0.72cm]{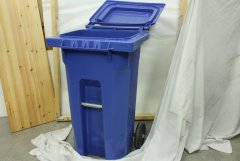}} & Recycle & 2864$\times$1924 & 260 & 37 & 102 & 203 & 220 & 220 & 1.34--2.53 & 5.2 & 2.5 \\
\raisebox{-0.45\height}{\includegraphics[height=0.72cm]{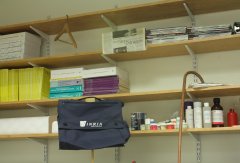}} & Shelves & 2952$\times$2000 & 240 & 54 & 113 & 192 & 200 & 200 & 2.53--4.14 & 4.2 & 2.8 \\
\raisebox{-0.45\height}{\includegraphics[height=0.72cm]{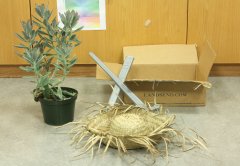}} & Sword1 & 2912$\times$2020 & 260 & 5 & 95 & 204 & 233 & 233 & 2.28--4.02 & 7.3 & 3.0 \\
\raisebox{-0.45\height}{\includegraphics[height=0.72cm]{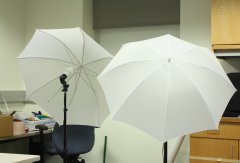}} & Umbrella & 2960$\times$2016 & 250 & 43 & 113 & 210 & 218 & 212 & 3.04--6.44 & 5.1 & 2.5 \\
\bottomrule
\end{tabular}
}
\end{table}

\section{Full results}
\label{app:results}
\begin{table*}[p]
\centering
\caption{Accuracy (non-occluded ground truth only) and cost per frame on Jetson AGX Orin (MAXN).
Dens.: fraction of non-occluded ground-truth pixels with a prediction. bad-2 counts missing predictions
as errors; AbsRel and $\delta_{1.05}$ are over predicted pixels. AbsRel$^{\dagger}$: monocular output
after per-image oracle median scaling. GPU ms: median TensorRT GPU compute time for networks and median
Nsight Systems GPU-busy time for libSGM. J/frame: dynamic SoC energy. Accuracy of Depth Pro, Depth Anything V2-S at full resolution,
Depth Anything~3, MoGe-2, Metric3D v2 and UniDepth V2 is from FP32 PyTorch and their timing from the
FP16 engines (\cref{sec:fp16}).}
\label{tab:main}
\small
\adjustbox{max width=\textwidth, max totalheight=0.88\textheight}{\begin{tabular}{llrrrrrrr}
\toprule
Method & Resolution & Dens. & bad-2 & AbsRel & $\delta_{1.05}$ & AbsRel$^{\dagger}$ & GPU ms & J/frame \\
 & & [\%] & [\%] & [\%] & [\%] & [\%] & & \\
\midrule
Light ESS (NVIDIA) & 288$\times$480 (fixed) & 100 & 36.9 & 1.16 & 96.7 & -- & 2.8 & 0.08 \\
libSGM & $\approx$497$\times$720 (quarter) & 88 & 31.3 & 1.18 & 94.6 & -- & 6.1 & 0.18 \\
libSGM (hole-filled) & $\approx$497$\times$720 (quarter) & 100 & 28.4 & 2.13 & 90.3 & -- & 6.1 & 0.18 \\
ESS (NVIDIA) & 576$\times$960 (fixed) & 100 & 20.4 & 0.70 & 98.4 & -- & 8.6 & 0.31 \\
libSGM & $\approx$995$\times$1440 (half) & 85 & 24.7 & 0.73 & 96.8 & -- & 25.1 & 0.73 \\
libSGM (hole-filled) & $\approx$995$\times$1440 (half) & 100 & 20.2 & 1.79 & 92.1 & -- & 25.1 & 0.73 \\
Fast-FoundationStereo 20-30-48 & 384$\times$576 & 100 & 17.0 & 0.64 & 98.2 & -- & 67.7 & 1.88 \\
libSGM & full ($\approx$1990$\times$2880) & 76 & 32.1 & 0.63 & 97.5 & -- & 100.6 & 2.96 \\
libSGM (hole-filled) & full ($\approx$1990$\times$2880) & 100 & 22.3 & 1.92 & 91.4 & -- & 100.6 & 2.96 \\
Fast-FoundationStereo 23-36-37 & 384$\times$576 & 100 & 12.7 & 0.51 & 98.7 & -- & 115.6 & 3.32 \\
C-Fast-FoundationStereo & 384$\times$576 & 100 & 12.9 & 0.52 & 98.7 & -- & 115.7 & 3.29 \\
Fast-FoundationStereo 20-30-48 & 512$\times$768 & 100 & 12.4 & 0.51 & 98.7 & -- & 149.7 & 4.23 \\
Fast-FoundationStereo 23-36-37 & 512$\times$768 & 100 & 8.9 & 0.42 & 99.0 & -- & 257.5 & 7.43 \\
C-Fast-FoundationStereo & 512$\times$768 & 100 & 9.4 & 0.43 & 99.0 & -- & 257.6 & 7.39 \\
Fast-FoundationStereo 20-30-48 & 768$\times$1152 (max) & 100 & 8.5 & 0.40 & 99.0 & -- & 444.7 & 12.93 \\
Fast-FoundationStereo 23-36-37 & 768$\times$1152 (max) & 100 & 6.5 & 0.33 & 99.3 & -- & 758.1 & 22.39 \\
C-Fast-FoundationStereo & 768$\times$1152 (max) & 100 & 6.0 & 0.33 & 99.3 & -- & 760.4 & 21.95 \\
\midrule
Depth Anything V2-S (native) & 518$\times$770 (native) & 100 & 98.1 & 29.88 & 11.0 & 8.93 & 11.5 & 0.50 \\
Depth Anything V2-B (native) & 518$\times$770 (native) & 100 & 98.5 & 30.55 & 7.3 & 8.93 & 26.8 & 1.32 \\
Depth Anything 3 Metric-L (GT focal) & 350$\times$504 (native) & 93 & 95.0 & 12.34 & 30.8 & 5.97 & 33.2 & 1.69 \\
UniDepth V2-S (GT intrinsics) & 644$\times$938 (native) & 93 & 98.4 & 24.67 & 11.9 & 5.71 & 42.5 & 1.72 \\
Metric3D v2-S (GT focal) & 616$\times$1064 (native) & 65 & 98.8 & 60.06 & 5.5 & 6.51 & 54.5 & 2.01 \\
MoGe-2-S (GT focal) & 700$\times$1008 (native) & 76 & 98.5 & 33.26 & 7.5 & 4.86 & 72.6 & 2.77 \\
MoGe-2-S (est. focal) & 700$\times$1008 (native) & 100 & 97.7 & 30.33 & 10.7 & 9.06 & 72.6 & 2.77 \\
Depth Anything V2-S & 1008$\times$1512 & 100 & 99.9 & 43.79 & 0.5 & 9.17 & 76.6 & 3.40 \\
Depth Anything V2-L (native) & 518$\times$770 (native) & 100 & 98.5 & 32.46 & 9.3 & 8.31 & 84.5 & 4.12 \\
UniDepth V2-B (GT intrinsics) & 644$\times$938 (native) & 84 & 98.6 & 25.99 & 11.6 & 3.90 & 85.0 & 3.69 \\
MoGe-2-B (GT focal) & 700$\times$1008 (native) & 96 & 97.5 & 21.58 & 17.5 & 3.88 & 102.6 & 4.19 \\
MoGe-2-B (est. focal) & 700$\times$1008 (native) & 100 & 99.7 & 36.41 & 1.5 & 7.53 & 102.6 & 4.19 \\
Depth Anything 3 Metric-L (GT focal) & 700$\times$1008 & 90 & 96.8 & 14.21 & 23.0 & 6.07 & 181.3 & 8.56 \\
UniDepth V2-L (GT intrinsics) & 644$\times$938 (native) & 99 & 96.4 & 20.47 & 12.5 & 4.09 & 182.1 & 8.28 \\
MoGe-2-L (GT focal) & 700$\times$1008 (native) & 94 & 95.8 & 19.24 & 13.0 & 3.78 & 211.6 & 9.28 \\
MoGe-2-L (est. focal) & 700$\times$1008 (native) & 100 & 98.6 & 39.87 & 6.9 & 9.01 & 211.6 & 9.28 \\
Metric3D v2-L (GT focal) & 616$\times$1064 (native) & 63 & 100.0 & 56.93 & 0.0 & 4.00 & 265.3 & 11.97 \\
UniDepth V2-S (GT intrinsics) & 1302$\times$1890 & 95 & 97.5 & 23.37 & 10.2 & 8.02 & 279.1 & 12.22 \\
Metric3D v2-S (GT focal) & 1232$\times$2128 & 14 & 100.0 & 133.00 & 0.0 & 7.94 & 295.6 & 12.11 \\
MoGe-2-S (GT focal) & 1400$\times$2030 & 68 & 98.5 & 42.94 & 7.2 & 5.60 & 418.7 & 16.71 \\
MoGe-2-S (est. focal) & 1400$\times$2030 & 100 & 99.8 & 36.67 & 1.7 & 9.75 & 418.7 & 16.71 \\
Depth Anything V2-L & 1008$\times$1512 & 100 & 99.6 & 39.46 & 3.6 & 9.92 & 502.7 & 23.54 \\
UniDepth V2-B (GT intrinsics) & 1302$\times$1890 & 95 & 96.8 & 21.21 & 19.0 & 5.83 & 579.6 & 25.47 \\
MoGe-2-B (GT focal) & 1400$\times$2030 & 86 & 99.1 & 34.36 & 8.2 & 4.43 & 663.4 & 28.11 \\
MoGe-2-B (est. focal) & 1400$\times$2030 & 100 & 99.7 & 37.07 & 4.2 & 8.26 & 663.4 & 28.11 \\
Depth Anything V2-S & 1988$\times$2884 (full) & 100 & 100.0 & 56.04 & 0.1 & 9.73 & 761.6 & 31.70 \\
Depth Pro (GT focal) & 1536$\times$1536 (fixed) & 98 & 95.9 & 11.98 & 28.9 & 4.70 & 802.5 & 38.68 \\
Depth Pro (est. focal) & 1536$\times$1536 (fixed) & 95 & 99.4 & 18.95 & 10.5 & 4.70 & 802.5 & 38.68 \\
Metric3D v2-S (GT focal) & 1988$\times$2884 (full) & 5 & 100.0 & 162.46 & 0.0 & 8.44 & 992.3 & 41.69 \\
UniDepth V2-S (GT intrinsics) & 1988$\times$2884 (full) & 96 & 98.2 & 22.57 & 10.1 & 8.96 & 1,133.3 & 50.43 \\
MoGe-2-S (GT focal) & 1988$\times$2884 (full) & 66 & 94.8 & 40.23 & 12.5 & 6.00 & 1,178.6 & 48.76 \\
MoGe-2-S (est. focal) & 1988$\times$2884 (full) & 100 & 99.3 & 41.33 & 2.6 & 9.99 & 1,178.6 & 48.76 \\
UniDepth V2-L (GT intrinsics) & 1302$\times$1890 & 100 & 98.3 & 25.02 & 8.3 & 5.55 & 1,250.1 & 56.50 \\
MoGe-2-L (GT focal) & 1400$\times$2030 & 87 & 97.1 & 23.36 & 13.9 & 4.05 & 1,503.3 & 67.23 \\
MoGe-2-L (est. focal) & 1400$\times$2030 & 100 & 99.1 & 40.65 & 5.1 & 9.33 & 1,503.3 & 67.23 \\
Metric3D v2-L (GT focal) & 1232$\times$2128 & 9 & 100.0 & 127.52 & 0.0 & 5.61 & 1,558.6 & 70.16 \\
Depth Anything V2-B & 1988$\times$2884 (full) & 100 & 100.0 & 41.24 & 0.5 & 10.64 & 1,628.4 & 72.17 \\
MoGe-2-B (GT focal) & 1988$\times$2884 (full) & 81 & 99.0 & 37.23 & 5.3 & 4.54 & 2,007.0 & 86.64 \\
MoGe-2-B (est. focal) & 1988$\times$2884 (full) & 100 & 99.6 & 36.32 & 3.3 & 8.35 & 2,007.0 & 86.64 \\
UniDepth V2-B (GT intrinsics) & 1988$\times$2884 (full) & 96 & 99.4 & 26.18 & 8.4 & 7.69 & 2,335.2 & 103.84 \\
Depth Anything V2-L & 1988$\times$2884 (full) & 100 & 99.7 & 42.69 & 2.3 & 10.72 & 4,496.9 & 202.70 \\
Depth Anything 3 Metric-L (GT focal) & 1988$\times$2884 (full) & 56 & 100.0 & 54.88 & 0.2 & 6.24 & 4,503.7 & 201.72 \\
MoGe-2-L (GT focal) & 1988$\times$2884 (full) & 86 & 99.1 & 24.45 & 13.1 & 3.99 & 4,817.9 & 214.81 \\
MoGe-2-L (est. focal) & 1988$\times$2884 (full) & 100 & 98.6 & 39.67 & 7.7 & 9.17 & 4,817.9 & 214.81 \\
UniDepth V2-L (GT intrinsics) & 1988$\times$2884 (full) & 100 & 100.0 & 28.10 & 2.9 & 6.48 & 5,047.5 & 226.71 \\
Metric3D v2-L (GT focal) & 1988$\times$2884 (full) & 8 & 100.0 & 136.76 & 0.0 & 5.23 & 5,297.5 & 235.45 \\
\bottomrule
\end{tabular}
}
\end{table*}

\clearpage
\section{libSGM parameters}
\label{app:libsgm}
\Cref{tab:libsgm} lists the libSGM parameters we use. $P_1$ and $P_2$ were chosen separately for each
input resolution by a grid search ($P_1\in\{1,2,3,4,6,10,15,20,30\}$, $P_2\in\{40,60,90,120,160,200,224\}$,
$P_1<P_2$) that minimises the mean AbsRel over the 10 evaluation scenes, scored exactly as in
\cref{tab:main}. All other parameters are libSGM's defaults, except that we use four aggregation paths
instead of eight and enable sub-pixel output. libSGM stores aggregated path costs in 8 bits, so the
census cost (at most 31 for the symmetric $9\times7$ census) plus $P_2$ must stay below 256; the
largest usable $P_2$ is therefore 224, and the search selected it at every resolution.

The search has no separate training set, so the selected parameters are tuned on the evaluation
scenes, which favours libSGM. To estimate how much, we repeated the selection ten times with one scene
held out and scored each held-out scene with the parameters chosen on the other nine. The held-out
mean AbsRel is 0.65\%, 0.75\% and 1.26\% at full, half and quarter resolution, against 0.63\%, 0.73\%
and 1.18\% when tuned on all ten scenes and 1.03\%, 1.04\% and 1.29\% with the default $P_1=10$,
$P_2=120$. The held-out choices agree with the selected ones in 9 of 10 folds at full and half
resolution and in 7 of 10 at quarter resolution. Minimising AbsRel over predicted pixels also lowers
the density: the stronger smoothing makes more pixels fail the left--right check (76\%, 85\% and 88\%
density, against 84\%, 89\% and 91\% with the defaults).

\begin{table}[!h]
\centering
\caption{libSGM (version 3.1.0, commit \texttt{e4c669b}) parameters.}
\label{tab:libsgm}
\small
\resizebox{\linewidth}{!}{%
\begin{tabular}{@{}lll@{}}
\toprule
Parameter & Value & Note \\
\midrule
$P_1$ & 4 (full), 6 (half), 3 (quarter) & tuned; default 10 \\
$P_2$ & 224 at every resolution & tuned; default 120; 224 is the 8-bit maximum \\
Aggregation paths & 4 (horizontal and vertical) & default 8 \\
Sub-pixel output & on (4 fractional bits) & default off \\
Disparity levels & 256 & maximum supported \\
Minimum disparity & 0 & default \\
Census & symmetric $9\times7$ (31 bits) & default \\
Uniqueness & 0.95 & default \\
Left--right check tolerance & 1\,px & default \\
Median filter & $3\times3$ & always applied \\
Input & 8-bit grey, area-downsampled for half and quarter & \\
Output & 16-bit, host-to-host (copies included in timing) & \\
\bottomrule
\end{tabular}}
\end{table}

\end{document}